\pdfoutput=1
\documentclass[doublespacing]{utdthesis}
\usepackage{times}
\usepackage{microtype}
\usepackage{amsmath,amssymb,amsthm}
\usepackage{graphicx}
\usepackage{url}
\usepackage[authoryear]{natbib}
\usepackage{bibentry}
\nobibliography*
\let\cite=\citep

\usepackage{rotating}
\usepackage{ifpdf}
\ifpdf
  \usepackage{hyperref}
\fi
\usepackage{enumitem}

\usepackage{bold-extra}

\usepackage{import}
\usepackage{setspace}

\usepackage{multirow}

\usepackage{amsmath}
\usepackage{wrapfig}
\usepackage{array,etoolbox}
\usepackage{color, colortbl}

\usepackage{booktabs}
\usepackage{makecell}

\newcounter{PseudocodeCounterDC}

\newcounter{PseudocodeCounterMI}

\newcounter{AlgorithmCounterDC}

\newcounter{AlgorithmCounterMI}

\definecolor{color:1}{rgb}{0.961,0.941,0.839}
\definecolor{color:2}{rgb}{0.718,0.839,0.757}
\definecolor{color:3}{rgb}{0.757,0.737,0.859}
\definecolor{color:4}{rgb}{0.941,0.82,0.839}
\definecolor{color:5}{rgb}{0.859,0.941,0.98}
\definecolor{color:6}{rgb}{0.898,0.780,0.878}

\usepackage[dvipsnames]{xcolor}

\usepackage[T1]{fontenc}

\author{Alex Treacher, Kevin Nguyen, Dylan Owens, Daniel Heitjan, Albert Montillo}

\title{UQ-ARMED: Uncertainty quantification of adversarially-regularized mixed effects deep learning for clustered (non-\textit{iid}) data}

\usepackage{caption}
\begin{document}

\maketitle

\mainmatter

%%%%%%%%%%%%%%%%%%%%%%%%%%%%%%%%%%%%%%%%%%%%%%%%%%%%%%%%%%%%%%%%%%%%%%%%%%%%%%%%%%%%%%%%%%%%%%%%%%%%
% b-ARMED
%%%%%%%%%%%%%%%%%%%%%%%%%%%%%%%%%%%%%%%%%%%%%%%%%%%%%%%%%%%%%%%%%%%%%%%%%%%%%%%%%%%%%%%%%%%%%%%%%%%%

\chapter*{Abstract}
This work demonstrates the ability to produce readily interpretable statistical metrics for model fit, fixed effects covariance coefficients, and prediction confidence. Importantly, this work compares 4 suitable and commonly applied epistemic UQ approaches, BNN, SWAG, MC dropout, and ensemble approaches in their ability to calculate these statistical metrics for the ARMED MEDL models. In our experiment, not only do the UQ methods provide these benefits, but several UQ methods maintain the high performance of the original ARMED method, some even provide a modest (but not statistically significant) performance improvement. The ensemble models, especially the ensemble method with a 90\% subsampling, performed well across all metrics we tested with (1) high performance that was comparable to the non-UQ ARMED model, (2) properly deweights the confounds probes and assigns them statistically insignificant p-values, (3) attains relatively high calibration of the output prediction confidence. The MC dropout models showed the lowest performance, and failed to provide non-statistically significant fixed effects covariate coefficients. The SWAG model’s performance was dependent on the learning rate. Specifically, the models with low learning rate underestimated the fixed effect covariate coefficient uncertainty providing very small standard errors causing very significant p-values for all covariate coefficients including for the synthetic probes. Lastly, The BNNs also performed reasonably well, showing good model performance, and almost achieved providing statistically insignificant p-values for the synthetic probe covariate coefficients (depending on your cut off), but did not perform as well as the ensemble approaches for both covariate coefficient statistical significance, or model prediction confidence. The largest potential downside to the ensemble approaches is the increased training time, however as discussed in results, the balance between wall clock time and available computational resources could be achieved through parallelization. Additionally, in many instances the models inference time is more important than training time, for which the ensemble approaches are tied as the fastest of the UQ methods. Based on these results, the ensemble approaches, especially with a subsampling of 90\%, provided the best all-round performance for prediction and uncertainty estimation, and achieved our goals to provide statistical significance for model fit, statistical significance covariate coefficients, and confidence in prediction, while maintaining the baseline performance of MEDL using ARMED.

\chapter{Introduction}

Linear models and standard deep learning approaches assume the data is independent and identically distributed (\emph{iid}). In biomedical data, there is often a clustering of the input data. For example, samples in medical data  are often correlated: (e.g. samples from the same subject, recorded by the same observer, or obtained from the same clinic). In the later case, the clustering can be due to data acquired with slighlty different protocols or  medical devices (e.g. brand of magnetic resonance imaging device). This clustering of subjects causes the data to be non-\emph{iid} as subjects from the same grouping are typically more correlated then data from other other groupings. As illustrated in the Simpson’s paradox \cite{Wagner.1982} these confounds can lead to Type 1 and Type 2 errors. That is, covariates can be found to have an association with target where none exists (false positive) or vice versa (false negative). 

In the statistics community, such clustering is accounted for through the use of linear mixed effects (LME) models. The LME is defined as $y_s = \beta_0+\beta_1x_s+\mu_{0,j}+\mu_{1,j}x_s+\epsilon_s$, where $x_s$ is the covariate of the $s$-th subject, $y_s$ is the prediction target (dependent variable) for the s-th subjects, $\beta_0$ and $\beta_1$ are the fixed effects intercept and slope which apply to the whole population, $\mu_{0,j}$ and $\mu_{1,j}$ are the random effects intercept and slope for the $j$ cluster respectively, and $\epsilon$ is the error. LME models explicitly separate and quantify the fixed effects, i.e. population effects, using $\beta$, from the random effects, i.e. cluster specific effects, using $\mu$. Accounting for random effects by modeling the fixed and random effects separately helps mitigate confounds and decrease Type 1 and Type 2 errors. Additionally, LME models report results with statistically-meaningful characterization including the statistical significance of each covariate coefficient and for the overall multivariate model fit. Importantly, using standard cut-offs (e.g. alpha-criterion $\alpha=0.01$) this statistical characterization allows for a principled separation of significant and insignificant covariates (e.g. for the general population using the fixed effects covariate coefficient's significance). A primary limitation of LME models is that they are incapable of capturing non-linear relationships between the covariates and target. Additionally, while nonlinear mixed effects models (NLME) have been proposed they are not data driven and typically require specification (prior knowledge) of nonlinear relations (priors) often in the form of differential equations describing an underlying physical phenomena. In medical applications, such prior knowledge is typically unavailable.

Deep learning (DL) predictive models have achieved successes across all areas of life sciences. They are a universal approximator which can learn any non-linear association between predictors and target in a data driven way \cite{Hornik.1989}. Accordingly, to provide a data driven non-linear mixed effects solution, mixed effects deep learning (MEDL) models have been proposed to combine the strengths of DL with the confound mitigation of LMEs \cite{Nguyen.2022b, Simchoni.2021, Xiong.2019}. Notably, the Adversarially-Regularized MEDL (ARMED) framework \cite{Nguyen.2022b} was found to perform well relative to other MEDL approaches across many tasks. The ARMED framework employs a multi-module architecture to explicitly model both the fixed and random effects, quantify them separately, and combine them to provide a mixed effects prediction. Briefly, the ARMED framework consists of three elements. First, it adds an adversarial network module to testing the conventional models features for their ability to predict cluster membership. It acts as a negative feedback encouraging the conventional network to identify fixed effects that apply to the whole dataset. Second, it adds a random effects network to explicitly capture the random effects (RE), including the option to model the RE intercept, and a linear or non-linear RE slope. 
% MAYBE WE DONT NEED THIS DETAIL:  This network takes the cluster and the latent representation of the covariates from the conventional network as input. To ensure the random effects are modeled from a normal distribution similar to the linear mixed effects models, the random effects network is trained using Bayesian layers via variational inference and an evidence lower bound (ELBO) loss function. After training, the mean value of each weight distribution is used to provide deterministic random effects model weights. 
The third aspect is the addition of a cluster membership predictor (Z-predictor) allowing the full mixed effects model to be used for prediction on data from clusters unseen during training. 
% MAYBE WE DONT NEED THIS DETAIL:  The unseen site data is mapped to a weighted combination of seen sites which is used for inference on the unit. 
The ARMED framework attains, in a data driven way, a non-linear mixed effects association from covariates to targets. This is a critical advance over both the traditional statistical LME model and the statistical non-linear mixed effects models that require known partial or ordinary differential equations to govern the form of the non-linear mixed effects model.

% WOUND THIS WORK ADDRESSES
While the ARMED framework empowers deep learning for mixed effects modeling, yielding improved generalization, mitigating confounds and increasing model interpretability, there are several substantial limitations in ARMED curtailing their use in practical situations, such as medical applications. The first limitation is that ARMED does not provide a statistically meaningful measure of feature importance for each covariate. Second it does not provide a statistically meaningful measure of the overall model significance. A final limitation in the ARMED framework, is that it provides only a point estimate of the ME prediction, without any associated prediction confidence. We note that these limitations are common to other proposed MEDL solutions \cite{Simchoni.2021, Xiong.2019}.  
% and non-linear mixed effects models. 
Characterizing covariates and the overall model in a statistically meaningful fashion is essential for neural networks to become widely trusted among life scientists and clinicians accustomed to characterizations such as p-values and 95\% confidence intervals (CIs). Additionally, the characterization of model prediction confidence is critical so that users know when to trust the model. The need to address these key MEDL limitations motivates the this work. %enhancement of MEDL with these capabilities, since they have been shown to be highly successful and powerful approaches for predictive modeling. 

%% QQ: I think you should begin by saying that: UQ methods learn a distribution of models rather than a single model and this is achieved by learning a distribution of weights. This is most clear way to introduce them. 
Uncertainty quantification (UQ) methods learn a distribution of models rather than a single model, which is achieved by learning a distribution or multiple sets of model weights. Formally, a UQ DL model estimates a probability density function for the weights given the data: $p(w|D)$ where $D = {(x_1,y_1),(x_2,y_2)...(x_n,y_n)}$ and $n$ is the number of samples in the data. Specifically this captures epistemic (model) uncertainty. There are multiple methods to estimate model uncertainty for deep learning. In this work we evaluate the suitability of 4 of the most successful approaches. (1) \textit{Bayesian neural networks} (BNNs) directly estimate a distribution for each weight to estimate a posterior of the weights given the data $p(w|D)$ \cite{Haubmann.2020}, unlike standard networks that just used a point estimate.
%%[QQ: Why did you change from D --> X?  What is different btwn BNN and the opening sentence]
(2) \textit{Stochastic weight averaging Gaussian} (SWAG), approximates Bayesian inference by fitting a Gaussian to the first moment of the stochastic gradient decent (SGD) iterates. The distribution of the models weights are defined as a multivariate Gaussian, which can then be sampled from to estimate the model's posterior \cite{Maddox.2019}. (3) \textit{Ensemble methods} train multiple conventional models, each learning a single point estimate for each weight in the neural network. Each model is trained with a different perturbation of the training dataset or weight initialization during model training. The set of models is then used to approximate a posterior distribution at inference time \cite{Wilson.2020}. (4) Lastly, \textit{Monte Carlo (MC) dropout} which has traditionally been  used as a regularization method and can also be used at test time to mimic an ensemble method. For the Bayesian approximation, each network neuron is optionally selected for temporary exclusion from the model by sampling from a binomial distribution. The exclusion of neurons occurs both during training and inference to provide a prediction uncertainty \cite{Srivastava.2014, Gal.2016, Gal.2016b}. For all of these UQ methods, the probability distribution of the weights can then be employed to efficiently quantify the model uncertainty. To estimate the overall significance of the model, the traditional approach is permutation testing \cite{Combrisson.2015}. However permutation testing often requires retraining the complete model up to 10,000 times, depending on the desired alpha criterion, which is computationally intractable in  many problems. UQ via the methods described above mitigates this issue by allowing to directly estimate the significance of the overall multivariate model fit by exploiting the learned distribution of models.
%Ideally you cite a paper that says its difficult to use PT for neural networks because it is  requires retraining the model many times which is often prohibitively computational expensive.

By combining UQ with the state of the art mixed effected deep learning framework, ARMED \cite{Nguyen.2022b}, this work brings 3 new key capabilities to mixed effects deep learning. In particular it allows the development of MEDL models that not only capture fixed and random effects to mitigate type 1 and 2 errors but also: 1) produces a true probabilistic confidence in the models’s ME prediction, 2) calculates the statistical significance of the overall model: did it truly learn an association between covariates and target, or was it just happenstance?, and 3) estimates the effect size of each FE covariate, its statistical significance, and 95\% CI. We evaluate 4 different UQ methods for their suitability for the ARMED framework to achieve these aims, and make a concrete recommendation of which UQ method best attains the above objectives while maintaining or improving the benefits of the ARMED framework. To demonstrate this, the UQ-ARMED models are trained and evaluated on the classification for subjects with mild cognitive impairment (MCI) to convert to Alzheimer's disease (AD) or remain stable within 2 years from their baseline visit. The Alzheimer's Disease Neuroimaging Initiative data is used due to its large number of longitudinally tracked subjects, and useful covariates. The non-linearity of complex medical data, combined with the clustering of subjects within each site, makes this a suitable dataset to test mixed effects deep learning.
% QQ: to publish we will need another test case

\chapter{Methods}

Several methods have been proposed for mixed effects deep learning (MEDL). Recently it has been shown that ARMED outperforms comparables including MeNet, LMMNN, as well as alternative strategies including meta-learning, domain adversarial modes, and models that take the cluster membership as an additional covariate \cite{Nguyen.2022b}. Therefore we have chosen to ARMED as our base mixed effects deep learning framework, that we will enhance through the application of uncertainty quantification (UQ) methods. ARMED models the output as a point estimate, and as such does not intrinsically provide a principled uncertainty quantification in its predictions. 
% Note ARMED is not deterministic.. there is a random init, and random SGD...
While the output from a softmax activation, a commonly used classification activation function, can be interpreted as a confidence, it has been shown to be inaccurate and in particular over estimates the confidence in the category with highest evidence \cite{Monarch.2021}. Additionally, ARMED does not provide covariate coefficient significance. There are numerous methods for integrating model (aka epistemic) uncertainty including: (1) Bayesian Neural Networks (BNN) trained via Laplace approximations (2) BNNs trained via Markov chain Monte Carlo, (3) BNNs trained via variational Inference, (4) Monte Carlo (MC) Dropout, (5) ensemble approaches, and (6) Stochastic Weight Averaging Gaussian (SWAG)\cite{Maddox.2019}. Laplace approximation, with its required calculation of the Hessian for complex architectures, often contains millions of parameters rendering it in practice to be computationally expensive, non-trivial, and error prone. Markov chain Monte Carlo (MCMC) requires integrating over the model weights of the model to estimate the prior and posterior distribution. This requires many model samples, many of which are discarded as burn in samples. These factors often make Laplace approximations MCMC intractable for large models \cite{Mena.2022, Abdullah.2022}. These methods were therefore excluded from this work. Instead, this work focuses on 4 UQ approaches including three practical and commonly employed approaches\cite{Abdullah.2022}, BNNs trained via variational inference (hence forth referred to as simply BNNs), MC dropout, and ensemble models. We also include SWAG which is a relatively newer but efficient approach \cite{Maddox.2019}). Collectively, the approaches considered are representative of the wide variety of methods proposed for UQ, enabling a comprehensive evaluation of the suitability of UQ for MEDL, specifically ARMED. 

\section{Uncertainty quantification ARMED models}
%M1
This work integrates UQ into the ARMED framework in order to obtain 3 objectives: (1) produce an accurate probabilistic confidence in the model’s ME prediction, allowing for the user to better determine when to trust the models prediction, (2) calculate the statistical significance of the overall model, to determine whether the model learned a genuine covariate to target association, (3) provide statistically-meaningful characterizations of each covariate including the effect size statistical significance and the 95\% confidence interval (CI).
% QQ: this seems redundant so I deleted it: also providing a statistical probability that there is a relationship between the covariate and target. 
% QQ: this seems to explain ARMED which is details described in that manu, which we can leave out. In ARMED to ensure the random effects are approximated using a normal distribution, the random effects model is trained as a BNN, with the fixed effects network being deterministic. 
% QQ: this was a bit unclear, and redundant with paragraph below, thus I endeavored to rewrite it below. To provide the UQ for the mixed effects model, while ensuring compatibility to the random effects network, the fixed effects network is adapted to model UQ. 

%In the original ARMED formulation is composed of two primary subnetworks: the fixed effects subnetwork and the random effects subnetwork. The random effects subnetwork is modeled with a BNN, i.e. its weights are drawn from a Gaussian, and variational inference is used to learn those weights. However for inference ARMED does not output a probabilistic confidence and instead collapses the distribution to the learned average weight to produce a deterministic point estimate model for predictions. In this work, we relax that constraint and we also embed the the Bayesian UQ method into the Fixed Effects subnetwork, so it too can learn a distribution of weights. In the original formulation the FE subnetwork learns only point estimates for each weight. 

For the overall model, hyperparameters are held consistent with the original ARMED network \cite{Nguyen.2022b}, however each UQ method has a different hyperparameter to optimize (e.g. for the MC dropout model, the dropout rate). To ensure that each of the 4 UQ methods is not hindered by a specific selection of the hyperparameter, 3-5 reasonable values for these hyperparameters are tested for each model, as described in sections \ref{sec:UQ-ARMED:methods:BNN}, \ref{sec:UQ-ARMED:methods:SWAG}, \ref{sec:UQ-ARMED:methods:Ensemble}, and \ref{sec:UQ-ARMED:methods:MC dropout}, and reported individually in the results. To provide a distribution of weights, the weight posterior is sampled 30 times for each UQ method. For the ensemble methods, this requires 30 models to be trained, but for the other UQ methods that allow for direct sampling from the posterior, a single model is trained. Note that in adding UQ to ARMED we do not expect performance to increase, rather we aim to achieve the above 3 practical goals \textit{while maintaining the performance} of the original ARMED approach.

\subsection{Bayesian Neural Networks for uncertainty quantification in mixed effects deep learning}
\label{sec:UQ-ARMED:methods:BNN}
% M2
For conventional networks with multiple layers, making any single layer Bayesian causes the training to learn a distribution of model weights, which is sufficient to calculate the model uncertainty. However, since we do not know a priori which subset of the layers to make Bayesian, this is treated as a hyperparameter in our experiments. To keep the number of models reasonable comparisons are made between models that have the first, last, or all layers Bayesian. Each layer selected to be Bayesian replaces each point estimate weight (scalar) with two values, a mean, and a variance that define Gaussian probability distribution for each weight. Other probability distributions are possible, however Gaussian distributions are typically used, unless there is prior knowledge that suggests a different distribution would perform better based on the dataset. As mentioned above, these models are trained via variational inference to maximize the Evidence Lower Bound (ELBO), $ ELBO = [log p(X|W)] - D_KL(q(W)||p(W)] $, that consists of the log likelihood and the KL divergence between the posterior and a prior distribution and $q(W)$ that defines a surrogate posterior approximating the true posterior. As our weights are modeled as a Gaussian, our prior distribution is also a Gaussian with 0 mean and unit variance.

\subsection{SWAG for uncertainty quantification in MEDL}
\label{sec:UQ-ARMED:methods:SWAG}
% M3
Similar to BNNs, SWAG is able to construct a distribution of model weights from training a single model, and thus also produces a posterior distribution of model weights that can be sampled from. However, rather than estimate each weight as a distribution like BNNs, SWAG estimates the variance/co-variance matrix for the models weights. To achieve this, after training until model convergence, SWAG samples the local minima by continuing to train with SGD and a constant learning rate to estimate the variance/co-variance of the weights within the local minima. There are two primary SWAG variations: the first, which is the full SWAG approach estimates the complete variance-covariance matrix, while the second estimates the diagonal variance matrix. For clarity, we refer to the approach which estimates the full variance-covariance matrix as SWAG-full, and the latter approach as SWAG-diag. We note that the SWAG-full approach explicitly estimates off diagonal covariance, which is not attained by other UQ methods such as BNN and thereby motivated to compare these UQ methods. For SWAG-full, the variance and variance co-variance matrix are scaled by half prior to summing as suggested in the original manuscript. This essentially provides the mean value between the estimated co-variance matrix, and the diagonal used in the SWAG-diag. Also, as described in the SWAG manuscript, the model is trained until convergence prior to estimating the weight variance and covariance matrices. For a fair comparison to the other methods, the model is trained to convergence using the same optimizer. An additional 30 training iterations with a consistent learning rate is then used to estimate the variance and co-variance matrices. For each SWAG model, the estimated variance and co-variance matrices depend on the learning rate as described in the original manuscript \cite{Maddox.2019}.  If the learning rate is too small not enough variance will be captured while sampling for the variance/co-variance matrix. If the learning rate is too large, the model may diverge from the local minima it is sampling from. Therefore, we test 4 different learning rates for comparison: 1e-1, 1e-2, 1e-3, and 1e-4 for each of the SWAG UQ methods. This set of learning rates were chosen to span multiple orders of magnitude commonly seen for learning rates, while simultaneously limiting the number models to train.

\subsection{Monte Carlo dropout for uncertainty quantification in mixed effects deep learning}
\label{sec:UQ-ARMED:methods:MC dropout}
% M4
MC dropout is a simpler approach to estimate model uncertainty when compared with BNNs and SWAG, but has also been shown to be an effective UQ method \cite{Srivastava.2014, Gal.2016, Gal.2016b}. To provide UQ, MC dropout randomly sets layer's input activation's to zero, and is often used as a regularization method while training. However, to estimate the model uncertainly multiple inferences are taken for each provided input, each time with different activation's being set to zero. MC dropout can be seen as an Bayesian estimate the models the weights, with the posterior modeled as a binomial distribution. The most important hyperparameter for MC dropout is the dropout rate that determines the number of input activation's to set to zero for each layer. Too large of a fraction, and the model does not have enough information to use for prediction, too small of a fraction and the model may underestimate the uncertainty. For our experiments we tested a dropout rate in increments of 0.1 from 0.1 to 0.5: 0.1, 0.2, 0.3, 0.4, and 0.5. This selection covers a large range of possible dropout rates, while maintaining a reasonable number of models to train and compare. 

\subsection{Ensemble approaches for uncertainty quantification in mixed effects deep learning}
\label{sec:UQ-ARMED:methods:Ensemble}
% M5
Unlike the preceding UQ methods which train only a single model to obtain a distribution of learned models, the ensemble UQ approach trains a set of deterministic models from different randomly chosen subsets of the data, or from different random initializations. After training, the distribution of conventional models can are used to make a distribution of predictions (one per conventional model) which subsequently allow for model uncertainty to be estimated. To ensure that the models are trained differently some perturbation on the training of the model is required. To provide this perturbation, we used (1) different model initializations, (2) sub-sampling of the training data with 70, 80, or 90\% of the data for the model to train on. This allows us to compare different ensemble methods (e.g. initializations, and sub-sampling). The probability distribution of the model may depend on the amount of sampling from the data, but we didn't want sample less than 70\% of the data as to not effect the final models performance. To enable fair comparison with the other UQ methods, a total of 30 conventional models are trained for each variety of ensemble methods.

\subsection{Network architecture and initialization}
% M6
For each experiment, all networks are trained consistently e.g., same optimizer and learning rate. The models configuration (a.k.a hyperparameters) are as consistent as possible, only modifying when necessarily for the specific Bayesian method with the ranges defined in the above sections: the BNN have additional weights to model the variance of the weights, SWAG trains for an additional 30 epochs post convergence with different learning rates, ensemble methods use different training perturbations, and MC dropout includes different dropout rates. Specific hyperparameters are included in supplemental table \ref{sup:UQ-ARMED:HP} and identical to the original ARMED manuscript \cite{Nguyen.2022b}. Adversarial networks, including the adversarial network in ARMED, can be hard to train and sometimes diverge \cite{Nowozin.2016}. Therefore, to provide the best comparison between the ARMED and the UQ-ARMED models we initialize the ARMED and all UQ-ARMED models with an equivalent set of weights for each fold that was found to converge well for the ARMED model. Some modifications are required for compatibility with the UQ method. For the BNN, the initial starting average of the distribution is set to the initial deterministic weight from the ARMED model. For the SWAG, MC dropout and sub-sampling ensemble models, the initial weights are the same as the ARMED model. However, for the random initializations UQ model it is not possible to set all initial weights to be the same, so each model is independently initialized.

\section{Data partitioning}
\label{sec:UQ-ARMED:meth:kfold}
% M7+ this + next 3 subsecs
To help provide an estimate of generalizable performance for deep learning models, data splitting for training and evaluation is essential. For the MCI conversion experiment, the training data and seen test data consists of the largest 20 study sites from ADNI, and the remaining 34 sites are held out for the un-seen test data. 10 fold cross validation was used to partition the 20 sites into a training and seen-test set. To allow a fair comparison against the original ARMED \cite{Nguyen.2022b} the same data partitioning strategy is used.

\section{Probe generation}
\label{sec:UQ-ARMED:methods:probe gen}
A fundamental goal of ARMED is to mitigate confounds. To test that this is being achieved, artificial covariates are added, as in the original ARMED manuscript. 5 probes are generated that are non-linearly associated with each site and the probability of conversion, but not associated with biological relevance. These probes are then added to the covariates to evaluate the ability for each network to provide non-significant statistical covariate coefficients for the 5 probes. 

\section{Performance metrics and methods for comparison between UQ methods.}
\label{sec:UQ-ARMED:methods:comparison metrics}
To compare the ability of the UQ methods to achieve our goals of the following subsections describe the comparison for: model predictive power  equivalent to ARMED and significance of model fit, ability to provide non-statistically significant fixed effects covariate coefficients to the probes, and calibration of model prediction, the following metrics are compared.

\subsection{Model Prediction Performance}
Standard performance metrics are used to compare overall predictive power between the standard ARMED and the Bayesian methods. These performance metrics include: AUROC Balanced accuracy (at Youden), AUROC, F1, F1 score, Spec at Youden, Sens at Youden. 

\subsection{Calculation of statistical significance for model fit}
\label{sec:UQ-ARMED:meth:modelfit}
% M8
At inference time, the models weights are sampled 30 times to produces a distribution of model balanced accuracy. To scale the balanced accuracy between –infinity and +infinity, which is ideal for statistical testing of accuracy, as it's range is limited to between 0-1, the $log(\frac{p}{1-p})$ transform is used. To provide a statistical significance for model fit, this transformed distribution is then compared to correspondingly scaled chance performance ($log(\frac{0.5}{1-0.5})=0$) using a 1 sided t-test where the null hypothesis is $H_0$: model performance $\leq$ to chance performance. We use a one sided test as we’re specifically interested in determining if the model’s performance is larger than chance. Rejecting this null hypothesis means that the model has learned a mapping of the input covariates to the target which is statistically significant for a given alpha criterion (false positive rate), where we define $\alpha$ = 0.05. While k-fold is a vital method in deep learning to minimize the generalization error, one wrinkle from k-fold cross validation approach is that this yields multiple models, each with their own statistical metrics. To remedy this, pooled statistical values across the models are calculated using the Satterthwaite approximation. The Satterthwaite approximation is used such that no assumption about the equality of the variances across folds needs to be made. The Satterthwaite provides an estimation of the standard error and degrees of freedom such that $SE = \sqrt{\sum_i^k{\frac{s_i^2}{n_i}}}$, in contrast to the un-pooled standard error $SE=\sqrt{\frac{s^2}{n}}$ where $s^2$ is the variance, n is the number of samples, and $i$ is the i-th sample. The Satterthwaite estimation for the degrees of freedom is calculated to be $df = \frac{(\sum_i^k\frac{s_i^2}{n_i})^2}{\sum_i^k(\frac{1}{n_i-1})(\frac{s_i^2}{n_i})^2}$. The pooled $SE$ and $df$ can then be used to calculate a pooled T and corresponding T value. 

\subsection{Calculation and comparison of statistical significance for covariate coefficients}
\label{sec:UQ-ARMED:methods:covariate coef}
% M9
The ARMED model is by definition a non-linear model. Therefore, estimating the covariate coefficient can be non-trivial. Gradient methods are a common approach to estimate the covariate coefficient for a deep learning model (TODO cite Grad-Cam and others?). ARMED employs a similar approach to estimate the mixed, fixed, and random, and effects covariate coefficient values. Specifically, we use: $\frac{\partial y_F}{\partial x}$, $\frac{\partial y_M}{\partial x}$, $\frac{\partial h_R}{\partial x}$, respectively. As each input matrix $X$ will provide p (number of samples) estimates for each covariate coefficient averaging is used to summarize these estimates into a single covariate measure. There are two reasonable approaches we could use, (1) take the gradient for each input vector, e.g. $ave(\frac{\partial y}{\partial x})$, (2) average the input vector and then take the gradient e.g. $\frac{\partial y}{\partial \Bar{x}}$. In the results we focus on (1), but also provide (2) in the supplemental. The benefits and limitations to each method is further discussed in the discussion section.

Sampling from the model’s weight posterior allows us to calculate a distribution of the covariate coefficients. Similarly to the model fit statistical significance, this allows us to compute a statistical significance for each covariate coefficient. Analogous to the linear mixed effects model, our null hypothesis is: $H_0: \beta = 0$ for each covariate coefficient, and a two sided t-test is used as we also want to account for covariate directionality. Similarly to the model fit section \ref{sec:UQ-ARMED:meth:modelfit}, we pool these statistics across the k folds via the Satterthwaite approximation.

\subsection{Model confidence}
\label{sec:UQ-ARMED:methods:confidence}
% M10
As the UQ methods provide a distribution of prediction, a confidence in that prediction can be calculated. To determine the validity of this confidence, a comparison between the confidence of the correctly and in-correctly predicted subjects is used. Well calibrated prediction confidence will show lower confidence on the subjects that are incorrectly predicted. The larger difference between the confidence of the correctly and incorrectly predicted subjects, the more accurate the model confidence.

The prediction confidence of the model is the defined as the integral over the prediction probability with respect to the prediction. This integral is taken either side of the decision boundary(s) (e.g. 0.5 for a 2 class classifier). The max value for the integral between each decision boundary is then used as the confidence. Sampling from the posterior provides a discrete distribution of predictions, and therefore a sum takes place of the integral, which is then normalized to the total number of samples. For example, if after producing 30 predictions for a binary classification problem, 3 predictions are for class 0, and 27 for class 1, then the confidence would by max(3/30, 27/30), or 90\% confidence for a prediction of class 1 \cite{}. To pool across folds for the seen sights, each Bayesian deep learning method provides a single confidence for each subject. Therefor there is no need to pool, rather a simple concatenation providing a confidence score for each subject in the seen test data is all that is needed. To pool across folds for the unseen sites each method will provide k (where k is the number of folds) confidences. To eliminate the possibility of averaging confidences across models that make different predictions, a joint distribution is calculated across k folds and a final confidence is calculated based on that joint probability distribution. 

\subsection{Train and inference time comparison}
A final practical consideration for these methods is the amount of computational resources they use. Therefore, the training and inference time for these models is recorded and compared. For comparison both training and inference time across the different ARMED methods is measured and summarized the across the k-folds.

\section{Experimental Materials}
\subsection{Alzheimer’s Disease Neuroimaging Initiative (ADNI) dataset}
\label{sec:UQ-ARMED:materials:ADNI}
% M11
The ADNI dataset is used to provide a real word application to test and compare the UQ-ARMED methods on. As described in sec \ref{sec:UQ-ARMED:results:MCI}, UQ-ARMED methods are compared for the prediction of if a subject with mild cognitive impairment (MCI) will convert to Alzheimer's disease (AD) within 24 months. The data included in this study in ADNI are are MCI subjects with baseline demographic information, cognitive scores, tabular summary metrics from neuroimaging measurements, and bio-marker measurements, with a follow up diagnosis in 24 months. This provides a total of 42 features. The follow up diagnosis provides the target for each subject. Subjects that convert to AD within 24 months are labeled as progressive MCI (pMCI), and the remaining subjects that do not convert are labeled as stable MCI (sMCI). Correlation between subjects from different sites can be caused from differences in imaging protocols and inter-rater variability. To explicitly account for the site confound, it is modeled as the random effect.

A total of 719 subjects from 54 sites match our requirements, 392 from the 20 sites with the largest number of subjects are included in the training and seen test (with k-fold applied), and the unseen test consists of the remaining 327 from 34 sites. 27.041\% of the subjects training and seen test are pMCI, and 37.615\% of the unseen subjects pMCI.

\chapter{Experimental Results}
\section{Application of UQ-ARMED to distinguish stable versus progressive MCI}
\label{sec:UQ-ARMED:results:MCI}
An mild cognitively impaired (MCI) patient is a subject with a cognitive disorder including memory and executive function issues which don’t yet impair daily activities. This experiment compares non-UQ ARMED (the original ARMED approach), and ARMED with 4 different UQ methods for the two-category classification task of predicting whether an MCI patient has stable (sMCI) or progressive (pMCI) MCI. A pMCI patient is a subject whose symptoms worsen within 24 months from baseline to full Alzheimer’s disease (AD) where cognitive issues impair daily activities. sMCI patients are subjects who retain stable in their MCI diagnosis at the 24 month follow up. The primary random effect in this cross-sectional dataset is from correlation among samples from the same site. There are 54 different sites in the data. Subjects from 20 sites are used for training (partitioned into training and test data from seen sites), and the remaining sites are held out as test data from unseen sites. As summarized in materials section  \ref{sec:UQ-ARMED:materials:ADNI}, the tabular summary data provided by ADNI (i.e. ‘ADNIMERGE’) are the 42 input covariates. A dense feed-forward neural network (AKA a multi layer perception (MLP)) was chosen as the conventional model, as this architecture is most suitable to learn from such tabular data. Comparable UQ ARMED models are constructed by adding the appropriate modification, as described in methods sections \ref{sec:UQ-ARMED:methods:BNN}, \ref{sec:UQ-ARMED:methods:SWAG}, \ref{sec:UQ-ARMED:methods:MC dropout}, and \ref{sec:UQ-ARMED:methods:Ensemble}, and evaluated as described in section \ref{sec:UQ-ARMED:methods:comparison metrics}, with hyperparameters defined in sup \ref{sup:UQ-ARMED:HP}. Across the models, we compare predictive performance, statistical significance for model fit, statistical significance of the covariates including the synthetic probes, and the calibration of the prediction confidence. 
% QQ: you need to say SOMETHING About the non-probe covariates. Are they reasonable? Consider for example ARMED manu. Is there evidence in the literature that these non-probe covariates are legitimate? In otherwords its insufficient to deweight the probes if the remaining covariates are not supported at all. We dont want a drastic decrease in the upweighting of unsupported non-probe covariates. I know, "there is no ground truth" but any attempt to tie to the literature is very important/ critical.

\section{Predictive Performance Comparison}
\label{sec:UQ-MABO:results:performance}
\begin{figure}[]
      
      \centering
      \includegraphics[width=\textwidth,keepaspectratio]{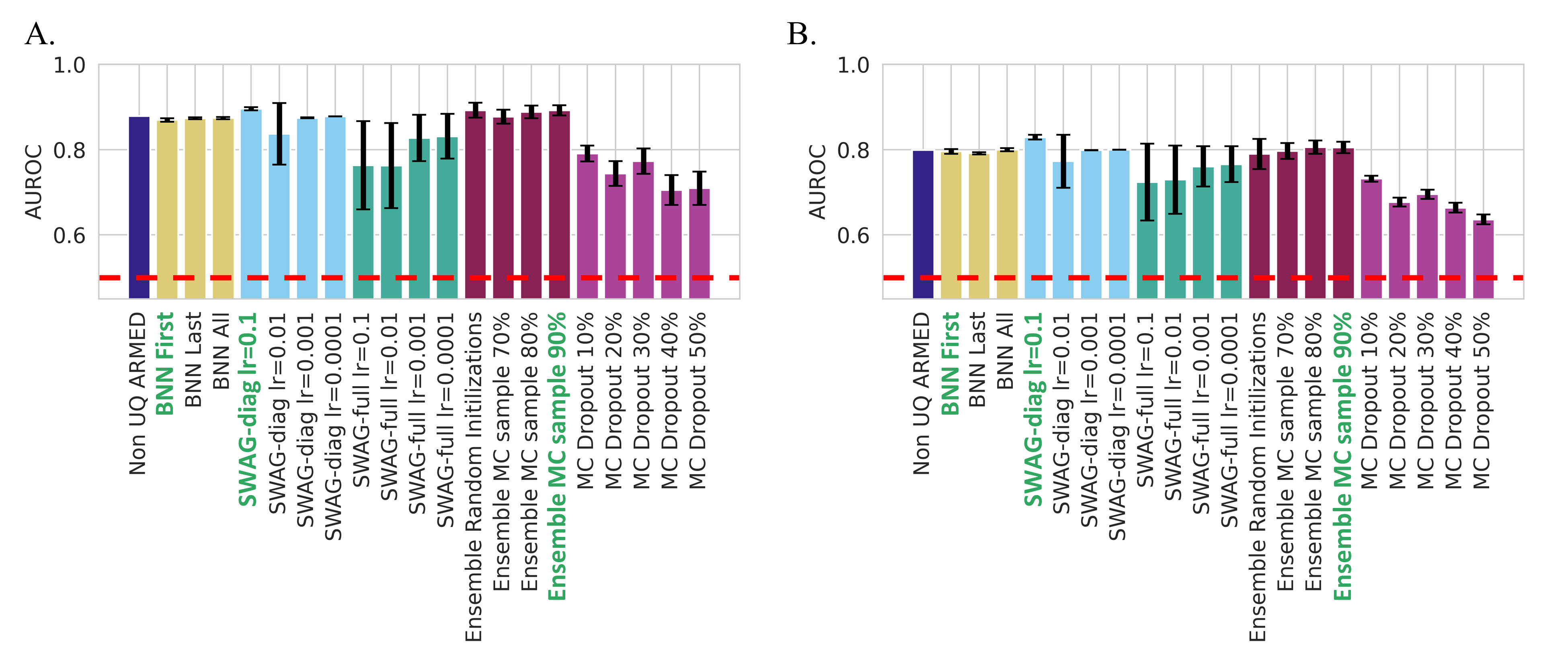}
      \caption[Comparison of AUROC performance of the tested UQ-ARMED methods.]{
       Comparison of AUROC performance of the tested UQ-ARMED methods, with error bars indicating the pooled standard error across the 30 draws from the model distribution. 
       Chance performance is displayed as a red dashed line. All models perform significantly better than chance. Exact AUROC values and 95\% CI are reported in table \ref{tab:UQ-ARMED:performance} under the AUROC column.
        \textbf{A.} AUROC on the seen-site test data. 
        \textbf{B.} AUROC on the unseen-site test data.
       Models highlighted by green text are focused on in subsequent main results figures, figures for other models found in supplementary materials. 
      }
      \label{fig:UQ-ARMED:performance}
    
    \end{figure}

As defined in the introduction, the goals of this work is to provide UQ for the models, while also maintaining the high performance archived by the non-UQ ARMED model. For this experiment, the non UQ ARMED model achieved an AUROC of 0.879 and 0.799 on the seen and unseen test data, respectively. Fig. \ref{fig:UQ-ARMED:performance} provides a comparison of AUROC across the models, and table \ref{tab:UQ-ARMED:performance} provides all  performance metrics for each model, including the 95\% CI for seen and unseen test data, and the overall model fit for the data calculated as described in \ref{sec:UQ-ARMED:meth:modelfit}. Based on the 95\% confidence interval (CI) of model performance, many UQ-ARMED methods (bolded in \ref{tab:UQ-ARMED:performance}) achieve the same (non-statistically different) or better performance when compared to non-UQ ARMED, fulfilling the performance objective.  Among UQ methods, we observe that all MC dropout models decreased performance significantly. Because of this, our remaining results analysis focuses on higher performing models, specifically: BNN All, Ensemble sampling with 0.9 fraction, and SWAG-diag with a learning rate=0.1. Results for lower performing UQ methods are included in the supplementary material \ref{appendix:UQ-ARMED}.

% QQ: Is this visualization supposed to match a subset (e.g. the first column of the preceding table?, is so state that in the figure caption. Also Never describe a plot by its shape "this bar plot is a bar plot" ... that adds no value.        
    
\begin{sidewaystable}[]
    \scriptsize
    \begin{tabular}{llllllllllllll}
                                                & p                    & \multicolumn{2}{l}{AUROC}                   & \multicolumn{2}{l}{Accuracy}                & \multicolumn{2}{l}{Spec. at Youden}         & \multicolumn{2}{l}{Sens. at Youden}         & \multicolumn{2}{l}{Sens. at 80\% Spec.}     & \multicolumn{2}{l}{Sens. at 90\% Spec.}     \\ \cline{2-14} 
    Model                                       &                      & mean                 & 95\% CI              & mean                 & 95\% CI              & mean                 & 95\% CI              & mean                 & 95\% CI              & mean                 & 95\% CI              & mean                 & 95\% CI              \\ \hline
    \multicolumn{1}{c}{Seen Site Performance}   & \multicolumn{1}{c}{} & \multicolumn{1}{c}{} & \multicolumn{1}{c}{} & \multicolumn{1}{c}{} & \multicolumn{1}{c}{} & \multicolumn{1}{c}{} & \multicolumn{1}{c}{} & \multicolumn{1}{c}{} & \multicolumn{1}{c}{} & \multicolumn{1}{c}{} & \multicolumn{1}{c}{} & \multicolumn{1}{c}{} & \multicolumn{1}{c}{} \\
    Non UQ ARMED                                &                      & \textbf{0.878}       & nan - nan            & 0.795                & nan - nan            & 0.846                & nan - nan            & 0.745                & nan - nan            & 0.780                & nan - nan            & 0.593                & nan - nan            \\
    BNN First                                   & \textless{}.0001     & 0.869                & 0.862 - 0.877        & \textbf{0.795}       & 0.780 - 0.810        & \textbf{0.817}       & 0.787 - 0.846        & \textbf{0.773}       & 0.750 - 0.796        & \textbf{0.777}       & 0.748 - 0.805        & \textbf{0.540}       & 0.477 - 0.602        \\
    BNN Last                                    & \textless{}.0001     & 0.873                & 0.870 - 0.877        & \textbf{0.806}       & 0.786 - 0.825        & \textbf{0.832}       & 0.817 - 0.848        & \textbf{0.779}       & 0.737 - 0.820        & 0.754                & 0.738 - 0.771        & \textbf{0.561}       & 0.525 - 0.597        \\
    BNN All                                     & \textless{}.0001     & \textbf{0.874}       & 0.869 - 0.879        & \textbf{0.793}       & 0.768 - 0.818        & 0.817                & 0.79 - 0.845         & \textbf{0.769}       & 0.719 - 0.818        & \textbf{0.759}       & 0.730 - 0.787        & \textbf{0.583}       & 0.539 - 0.627        \\
    SWAG-diag lr=0.1                            & \textless{}.0001     & \textbf{0.895}       & 0.888 - 0.903        & \textbf{0.801}       & 0.763 - 0.838        & \textbf{0.849}       & 0.793 - 0.904        & \textbf{0.753}       & 0.663 - 0.844        & \textbf{0.818}       & 0.766 - 0.869        & \textbf{0.586}       & 0.540 - 0.633        \\
    SWAG-diag lr=0.01                           & 0.0008               & \textbf{0.837}       & 0.696 - 0.978        & \textbf{0.769}       & 0.685 - 0.852        & \textbf{0.843}       & 0.693 - 0.992        & \textbf{0.695}       & 0.505 - 0.885        & \textbf{0.721}       & 0.543 - 0.900        & \textbf{0.537}       & 0.360 - 0.714        \\
    SWAG-diag lr=0.001                          & \textless{}.0001     & \textbf{0.874}       & 0.873 - 0.876        & \textbf{0.791}       & 0.776 - 0.805        & \textbf{0.841}       & 0.823 - 0.859        & \textbf{0.740}       & 0.701 - 0.779        & \textbf{0.763}       & 0.734 - 0.792        & \textbf{0.568}       & 0.554 - 0.582        \\
    SWAG-diag lr=0.0001                         & \textless{}.0001     & \textbf{0.878}       & 0.878 - 0.878        & \textbf{0.791}       & 0.791 - 0.791        & \textbf{0.846}       & 0.846 - 0.846        & \textbf{0.735}       & 0.735 - 0.735        & \textbf{0.780}       & 0.780 - 0.780        & \textbf{0.593}       & 0.593 - 0.593        \\
    SWAG-full lr=0.1                            & 0.5082               & \textbf{0.763}       & 0.560 - 0.967        & \textbf{0.663}       & 0.508 - 0.818        & \textbf{0.687}       & 0.354 - 1.020        & \textbf{0.638}       & 0.250 - 1.027        & \textbf{0.568}       & 0.255 - 0.882        & \textbf{0.356}       & 0.094 - 0.618        \\
    SWAG-full lr=0.01                           & 0.3556               & \textbf{0.763}       & 0.567 - 0.959        & \textbf{0.668}       & 0.548 - 0.789        & \textbf{0.733}       & 0.426 - 1.040        & \textbf{0.603}       & 0.251 - 0.955        & \textbf{0.552}       & 0.281 - 0.823        & \textbf{0.390}       & 0.142 - 0.639        \\
    SWAG-full lr=0.001                          & 0.0092               & \textbf{0.827}       & 0.721 - 0.934        & \textbf{0.738}       & 0.648 - 0.829        & \textbf{0.767}       & 0.562 - 0.973        & \textbf{0.709}       & 0.505 - 0.914        & \textbf{0.677}       & 0.504 - 0.850        & \textbf{0.504}       & 0.335 - 0.672        \\
    SWAG-full lr=0.0001                         & 0.0059               & \textbf{0.831}       & 0.728 - 0.934        & \textbf{0.745}       & 0.656 - 0.835        & \textbf{0.772}       & 0.584 - 0.961        & \textbf{0.719}       & 0.532 - 0.905        & \textbf{0.693}       & 0.526 - 0.861        & \textbf{0.509}       & 0.352 - 0.666        \\
    Ensemble Random Initializations              & \textless{}.0001     & \textbf{0.892}       & 0.858 - 0.927        & \textbf{0.788}       & 0.736 - 0.841        & \textbf{0.836}       & 0.767 - 0.905        & \textbf{0.741}       & 0.629 - 0.853        & \textbf{0.783}       & 0.678 - 0.888        & \textbf{0.640}       & 0.516 - 0.763        \\
    Ensemble MC sample 70\%                     & \textless{}.0001     & \textbf{0.877}       & 0.846 - 0.909        & \textbf{0.781}       & 0.727 - 0.835        & \textbf{0.836}       & 0.767 - 0.904        & \textbf{0.727}       & 0.605 - 0.849        & \textbf{0.764}       & 0.664 - 0.865        & \textbf{0.571}       & 0.440 - 0.702        \\
    Ensemble MC sample 80\%                     & \textless{}.0001     & \textbf{0.888}       & 0.859 - 0.918        & \textbf{0.792}       & 0.743 - 0.842        & \textbf{0.846}       & 0.781 - 0.912        & \textbf{0.739}       & 0.626 - 0.852        & \textbf{0.789}       & 0.689 - 0.889        & \textbf{0.590}       & 0.468 - 0.713        \\
    Ensemble MC sample 90\%                     & \textless{}.0001     & \textbf{0.892}       & 0.868 - 0.916        & \textbf{0.800}       & 0.752 - 0.849        & \textbf{0.852}       & 0.794 - 0.909        & \textbf{0.749}       & 0.644 - 0.855        & \textbf{0.798}       & 0.714 - 0.882        & \textbf{0.606}       & 0.496 - 0.715        \\
    MC Dropout 10\%                             & \textless{}.0001     & \textbf{0.791}       & 0.754 - 0.828        & 0.726                & 0.683 - 0.770        & \textbf{0.750}       & 0.684 - 0.816        & \textbf{0.703}       & 0.606 - 0.800        & \textbf{0.640}       & 0.544 - 0.736        & \textbf{0.413}       & 0.306 - 0.520        \\
    MC Dropout 20\%                             & 0.0018               & 0.744                & 0.687 - 0.801        & \textbf{0.703}       & 0.653 - 0.753        & \textbf{0.770}       & 0.698 - 0.841        & \textbf{0.637}       & 0.535 - 0.738        & \textbf{0.577}       & 0.453 - 0.701        & 0.365                & 0.249 - 0.481        \\
    MC Dropout 30\%                             & 0.001                & \textbf{0.773}       & 0.714 - 0.831        & \textbf{0.724}       & 0.667 - 0.782        & \textbf{0.751}       & 0.674 - 0.829        & \textbf{0.697}       & 0.579 - 0.816        & \textbf{0.594}       & 0.456 - 0.732        & \textbf{0.394}       & 0.283 - 0.505        \\
    MC Dropout 40\%                             & 0.1116               & \textbf{0.705}       & 0.636 - 0.774        & 0.668                & 0.606 - 0.73         & \textbf{0.695}       & 0.556 - 0.834        & \textbf{0.641}       & 0.498 - 0.784        & \textbf{0.520}       & 0.381 - 0.658        & \textbf{0.333}       & 0.214 - 0.451        \\
    MC Dropout 50\%                             & 0.0526               & \textbf{0.709}       & 0.633 - 0.786        & \textbf{0.681}       & 0.618 - 0.745        & \textbf{0.727}       & 0.649 - 0.805        & \textbf{0.636}       & 0.517 - 0.755        & \textbf{0.515}       & 0.365 - 0.664        & \textbf{0.310}       & 0.170 - 0.450        \\ \hline
    \multicolumn{1}{c}{Unseen Site Performance} & \multicolumn{1}{c}{} & \multicolumn{1}{c}{} & \multicolumn{1}{c}{} & \multicolumn{1}{c}{} & \multicolumn{1}{c}{} & \multicolumn{1}{c}{} & \multicolumn{1}{c}{} & \multicolumn{1}{c}{} & \multicolumn{1}{c}{} & \multicolumn{1}{c}{} & \multicolumn{1}{c}{} & \multicolumn{1}{c}{} & \multicolumn{1}{c}{} \\
    Non UQ ARMED                                &                      & 0.799                & nan - nan            & 0.713                & nan - nan            & 0.606                & nan - nan            & 0.820                & nan - nan            & 0.574                & nan - nan            & 0.398                & nan - nan            \\
    BNN First                                   & \textless{}.0001     & \textbf{0.796}       & 0.785 - 0.806        & \textbf{0.706}       & 0.695 - 0.717        & 0.574                & 0.552 - 0.596        & \textbf{0.838}       & 0.814 - 0.862        & \textbf{0.578}       & 0.550 - 0.605        & 0.395                & 0.363 - 0.427        \\
    BNN Last                                    & \textless{}.0001     & 0.791                & 0.786 - 0.796        & \textbf{0.704}       & 0.695 - 0.714        & \textbf{0.570}       & 0.548 - 0.591        & \textbf{0.839}       & 0.821 - 0.858        & \textbf{0.580}       & 0.568 - 0.593        & 0.406                & 0.385 - 0.426        \\
    BNN All                                     & \textless{}.0001     & \textbf{0.800}       & 0.793 - 0.806        & \textbf{0.707}       & 0.699 - 0.716        & 0.548                & 0.530 - 0.566        & \textbf{0.867}       & 0.851 - 0.884        & \textbf{0.586}       & 0.567 - 0.605        & 0.412                & 0.386 - 0.437        \\
    SWAG-diag lr=0.1                            & \textless{}.0001     & \textbf{0.829}       & 0.818 - 0.840        & \textbf{0.691}       & 0.665 - 0.718        & \textbf{0.666}       & 0.591 - 0.740        & 0.717                & 0.632 - 0.802        & \textbf{0.610}       & 0.545 - 0.675        & 0.454                & 0.397 - 0.511        \\
    SWAG-diag lr=0.01                           & 0.0127               & \textbf{0.773}       & 0.651 - 0.894        & \textbf{0.693}       & 0.633 - 0.752        & \textbf{0.630}       & 0.454 - 0.806        & \textbf{0.755}       & 0.550 - 0.959        & \textbf{0.568}       & 0.426 - 0.711        & 0.392                & 0.291 - 0.494        \\
    SWAG-diag lr=0.001                          & \textless{}.0001     & \textbf{0.799}       & 0.798 - 0.799        & \textbf{0.715}       & 0.711 - 0.719        & \textbf{0.624}       & 0.608 - 0.640        & \textbf{0.806}       & 0.793 - 0.819        & \textbf{0.576}       & 0.572 - 0.580        & 0.402                & 0.395 - 0.409        \\
    SWAG-diag lr=0.0001                         & \textless{}.0001     & \textbf{0.799}       & 0.799 - 0.799        & \textbf{0.714}       & 0.714 - 0.714        & \textbf{0.609}       & 0.609 - 0.609        & \textbf{0.819}       & 0.819 - 0.819        & \textbf{0.576}       & 0.576 - 0.576        & 0.398                & 0.398 - 0.398        \\
    SWAG-full lr=0.1                            & 1.3973               & \textbf{0.724}       & 0.546 - 0.901        & \textbf{0.588}       & 0.477 - 0.698        & \textbf{0.591}       & 0.201 - 0.980        & \textbf{0.585}       & 0.136 - 1.033        & \textbf{0.394}       & 0.112 - 0.677        & 0.258                & 0.041 - 0.475        \\
    SWAG-full lr=0.01                           & 0.9274               & \textbf{0.729}       & 0.573 - 0.886        & \textbf{0.622}       & 0.522 - 0.722        & \textbf{0.547}       & 0.208 - 0.886        & \textbf{0.697}       & 0.358 - 1.036        & \textbf{0.428}       & 0.196 - 0.660        & 0.291                & 0.119 - 0.463        \\
    SWAG-full lr=0.001                          & 0.1587               & \textbf{0.761}       & 0.667 - 0.854        & \textbf{0.667}       & 0.599 - 0.735        & \textbf{0.551}       & 0.295 - 0.807        & \textbf{0.783}       & 0.568 - 0.997        & \textbf{0.521}       & 0.367 - 0.675        & 0.354                & 0.230 - 0.479        \\
    SWAG-full lr=0.0001                         & 0.1146               & \textbf{0.766}       & 0.683 - 0.848        & \textbf{0.672}       & 0.606 - 0.738        & \textbf{0.553}       & 0.316 - 0.790        & \textbf{0.790}       & 0.589 - 0.991        & \textbf{0.531}       & 0.400 - 0.663        & 0.364                & 0.251 - 0.477        \\
    Ensemble Random Initializations              & 0.411                & \textbf{0.790}       & 0.720 - 0.859        & \textbf{0.653}       & 0.572 - 0.734        & \textbf{0.508}       & 0.212 - 0.804        & \textbf{0.798}       & 0.503 - 1.094        & \textbf{0.588}       & 0.465 - 0.712        & 0.393                & 0.280 - 0.507        \\
    Ensemble MC sample 70\%                     & 0.0128               & \textbf{0.797}       & 0.760 - 0.833        & \textbf{0.692}       & 0.636 - 0.748        & \textbf{0.565}       & 0.409 - 0.722        & \textbf{0.819}       & 0.675 - 0.963        & \textbf{0.592}       & 0.505 - 0.678        & 0.413                & 0.330 - 0.495        \\
    Ensemble MC sample 80\%                     & 0.0003               & \textbf{0.806}       & 0.775 - 0.837        & \textbf{0.708}       & 0.663 - 0.754        & \textbf{0.591}       & 0.467 - 0.715        & \textbf{0.826}       & 0.692 - 0.960        & \textbf{0.607}       & 0.533 - 0.682        & 0.428                & 0.357 - 0.500        \\
    Ensemble MC sample 90\%                     & 0.0002               & \textbf{0.805}       & 0.779 - 0.832        & \textbf{0.706}       & 0.663 - 0.749        & \textbf{0.605}       & 0.491 - 0.718        & \textbf{0.808}       & 0.682 - 0.934        & \textbf{0.606}       & 0.534 - 0.677        & 0.426                & 0.355 - 0.498        \\
    MC Dropout 10\%                             & 0.0173               & \textbf{0.732}       & 0.717 - 0.746        & \textbf{0.643}       & 0.623 - 0.663        & \textbf{0.533}       & 0.485 - 0.582        & \textbf{0.753}       & 0.697 - 0.808        & \textbf{0.470}       & 0.431 - 0.509        & 0.309                & 0.271 - 0.347        \\
    MC Dropout 20\%                             & 1.1758               & \textbf{0.677}       & 0.657 - 0.697        & \textbf{0.616}       & 0.596 - 0.636        & \textbf{0.569}       & 0.523 - 0.615        & 0.664                & 0.612 - 0.716        & \textbf{0.406}       & 0.368 - 0.444        & 0.238                & 0.200 - 0.275        \\
    MC Dropout 30\%                             & 0.9628               & \textbf{0.695}       & 0.674 - 0.716        & \textbf{0.620}       & 0.597 - 0.642        & \textbf{0.547}       & 0.494 - 0.600        & \textbf{0.693}       & 0.634 - 0.752        & \textbf{0.418}       & 0.376 - 0.460        & 0.267                & 0.227 - 0.306        \\
    MC Dropout 40\%                             & 1.9743               & \textbf{0.664}       & 0.641 - 0.686        & \textbf{0.584}       & 0.552 - 0.616        & \textbf{0.539}       & 0.453 - 0.626        & 0.629                & 0.512 - 0.745        & \textbf{0.351}       & 0.309 - 0.393        & 0.185                & 0.146 - 0.225        \\
    MC Dropout 50\%                             & 1.9964               & \textbf{0.636}       & 0.613 - 0.660        & \textbf{0.587}       & 0.567 - 0.608        & \textbf{0.540}       & 0.492 - 0.587        & \textbf{0.635}       & 0.580 - 0.689        & \textbf{0.349}       & 0.314 - 0.384        & 0.203                & 0.160 - 0.245       
    \end{tabular}
    
    \caption[Performance comparison for the prediction of stable vs. progressive mild cognitive impairment.]{
    Performance comparison for the prediction of stable vs. progressive mild cognitive impairment. P-values and confidence intervals were computed using pooled performance, sec \ref{sec:UQ-ARMED:meth:modelfit}. Models who's 95\% CI performance included or exceeded the non-UQ ARMED model are bolded.
    }
    
    \label{tab:UQ-ARMED:performance}
\end{sidewaystable}

\section{Statistical significance for covariate coefficient}
\label{sec:UQ-MABO:results:covariate}
The previous section focused on the performance and performance uncertainty, this section focuses on the statistical significance of the estimated co-variate coefficients for the fixed effects model. Table \ref{tab:UQ-ARMED:covariate summary} provides summary values for each of the models. These summary metrics show the p value, absolute value, and the rank for the fixed effect of the covariate of the most sadistically significant and highest rank synthetic probes. As the probes calculated to be random effects, high performing models should not provide statistically significant p-values. Overall the ensemble methods successfully deweighted the probes. Specifically, the ensemble with a 0.9 sampling model's highest ranked probe was ranked the 25th most important feature, and also assigned the probes insignificant p-values (p-value > 0.05), and not trending (p-value > 0.1) (table \ref{tab:UQ-ARMED:covariate summary}). The BNN methods also deweighted the probes, however is provided one of them (20\% of them) a statistically trending p-value (0.0669). Some SWAG models successfully deweighted the probes and assigned them insignificant p-values, e.g. SWAG-diag lr=0.01. While other SWAG models did not deweight the probes and assigned them statistically significant p-values. Lastly, the MC dropout models failed to deweight the probes and assigned them statistically significant p-values. 

    \begin{figure}[]
      
      \centering
      \includegraphics[width=\textwidth,keepaspectratio]{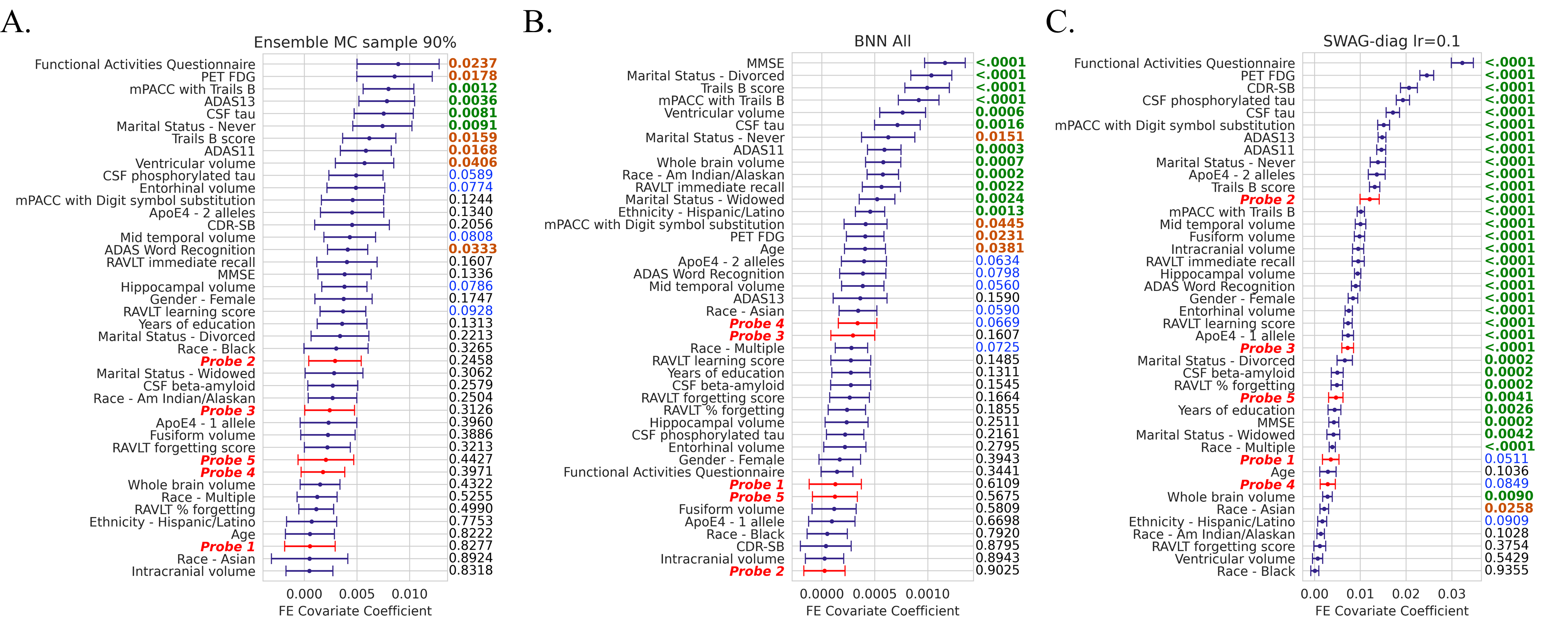}
      \caption[Coefficient covariate estimation and calculated uncertainty for 3 models with high predictive performance.]{
      Coefficient covariate estimation and calculated uncertainty for 3 models with high predictive performance. Each plot shows the mean, and 95\% CI for each estimated covariate coefficient. statistical significance is provded on the left and color coded to green: highly significant (p $<$ .01), orange: significant (0.01 $<$ p $<$ 0.05), blue: trending to significance (0.05 $<$ p $<$ 0.1), and black: not significant (p > 0.1). Synthetic probes, defined in section \ref{sec:UQ-ARMED:methods:probe gen} are displayed in red, and all other biological covariates are black.
      \textbf{A.} Absolute value of the covariate coefficient for the Ensemble model with a 90\% sampling of the data.
      \textbf{B.} Absolute value of the covariate coefficient for the BNN model with all layers Bayesian.
      \textbf{C.} Absolute value of the covariate coefficient for the SWAG-diag model with a 0.1 learning rate.
      Models shown here, are highlighted in green in fig \ref{tab:UQ-ARMED:covariate summary}. 
      Coefficient covariate estimation for all other models can be found in \ref{sup:UQ-ARMED:covariate coef}.
      }
      \label{fig:UQ-ARMED:covariate coef}
    
    \end{figure}
      
% QQ: The color coding key for p-values is missing. Green ... mans what?  Blue means ? Orange .. black      
% QQ: probes are red, original biological covariates are black. There were 42 original covariates and you are showing the top 20?

\begin{sidewaystable}[]
    \centering
    \begin{tabular}{lllllllll}
                                            
                                               & \multicolumn{4}{c}{Probe with smallest p}    & \multicolumn{4}{c}{Probe with largest   coefficient} \\
    MEDL method                                & Probe \# & p                & Coef    & rank & Probe \#   & p                  & Coef      & rank   \\ \hline
    Non UQ ARMED                               &  NA      & NA               & NA      &      & 2          &    NA                & 0.0033    & 14     \\
    BNN First                                  & 4        & 0.0055           & -0.0018 & 17   & 4          & 0.0055             & -0.0018   & 17     \\
    BNN Last                                   & 4        & \textless{}.0001 & -0.0027 & 13   & 4          & \textless{}.0001   & -0.0027   & 13     \\
    \textcolor{ForestGreen}{\textbf{BNN All}}  & 4        & 0.0669           & -0.0003 & 22   & 4          & 0.0669             & -0.0003   & 22     \\
    \textcolor{ForestGreen}{SWAG-diag lr=0.1}  & 3        & \textless{}.0001 & 0.0072  & 24   & 2          & \textless{}.0001   & 0.012     & 12     \\
    \textbf{SWAG-diag lr=0.01}                 & 3        & 0.1981           & 0.0029  & 25   & 3          & 0.1981             & 0.0029    & 25     \\
    \textbf{SWAG-diag lr=0.001}                & 3        & \textless{}.0001 & 0.0029  & 17   & 3          & \textless{}.0001   & 0.0029    & 17     \\
    \textbf{SWAG-diag lr=0.0001}               & 4        & \textless{}.0001 & -0.0027 & 21   & 2          & \textless{}.0001   & 0.0033    & 15     \\
    \textbf{SWAG-full lr=0.1}                  & 2        & 0.5774           & 0.0068  & 11   & 2          & 0.5774             & 0.0068    & 11     \\
    \textbf{SWAG-full lr=0.01}                 & 4        & 0.5852           & -0.0041 & 12   & 2          & 0.6352             & 0.0044    & 9      \\
    \textbf{SWAG-full lr=0.001}                & 4        & 0.5737           & -0.0025 & 17   & 4          & 0.5737             & -0.0025   & 17     \\
    SWAG-full lr=0.0001                        & 4        & 0.4814           & -0.0026 & 11   & 4          & 0.4814             & -0.0026   & 11     \\
    \textbf{Ensemble Random Initializations}    & 2        & 0.7474           & 0.0018  & 23   & 2          & 0.7474             & 0.0018    & 23     \\
    \textbf{Ensemble MC sample 70\%}           & 2        & 0.1847           & 0.0031  & 22   & 2          & 0.1847             & 0.0031    & 22     \\
    \textbf{Ensemble MC sample 80\%}           & 2        & 0.1853           & 0.0031  & 25   & 2          & 0.1853             & 0.0031    & 25     \\
    \textcolor{ForestGreen}{\textbf{Ensemble MC sample 90\%}}        & 2        & 0.2458           & 0.0029  & 25   & 2          & 0.2458             & 0.0029    & 25     \\
    MC Dropout 10\%                            & 4        & \textless{}.0001 & -0.0024 & 11   & 4          & \textless{}.0001   & -0.0024   & 11     \\
    MC Dropout 20\%                            & 5        & 0.0003           & -0.0021 & 9    & 5          & 0.0003             & -0.0021   & 9      \\
    \textbf{MC Dropout 30\%}                   & 5        & 0.0119           & -0.0016 & 8    & 5          & 0.0119             & -0.0016   & 8      \\
    MC Dropout 40\%                            & 5        & 0.0503           & -0.0016 & 8    & 5          & 0.0503             & -0.0016   & 8      \\
    MC Dropout 50\%                            & 4        & 0.2498           & -0.001  & 8    & 5          & 0.2928             & -0.0011   & 7     
    \end{tabular}

    \caption[Summary of synthetic probe fixed effects covariate coefficient.]{
    Summary of synthetic probe fixed effects covariate coefficient. This table provides the statistical metrics for the fixed effects covariate coefficient for the probes with the smallest p, and largest coefficient. Rank is the location of the probe when ranked by the absolute value of the coefficient. Coef and p is covariate coefficient and p-value of the probe, respectively, as described in \ref{sec:UQ-ARMED:methods:covariate coef}. Bolded UQ methods provided non-statistically significant p-values for all probes, and ranked them as low, or lower than the non-UQ ARMED model. NA = not applicable. The models in green are displayed in fig \ref{fig:UQ-ARMED:covariate coef}.
    }
    \label{tab:UQ-ARMED:covariate summary}
    
\end{sidewaystable}
% QQ: Label the first column "MEDL method"

\section{Prediction confidence comparison}
Results section \ref{sec:UQ-MABO:results:performance} and \ref{sec:UQ-MABO:results:covariate} both focus on generating model statistics comparable to the linear mixed effects model. However, UQ methods also allow us to provide a prediction probability distribution, allowing for a prediction confidence to be calculated as described in section \ref{sec:UQ-ARMED:methods:confidence}. To compare the validity of the prediction confidence for each Bayesian deep learning method, we compare the confidence calibration scores. In particular if the model's prediction confidence is accurate, we expect that a models confidence should be higher for its correct predictions (correctly classifying a subject as sMCI vs pMCI) than its incorrect predictions. Calibration of prediction confidence is calculated and compared for each model, as described in section \ref{sec:UQ-MABO:results:covariate}.

The prediction confidence for 3 top performing models is provided in \ref{fig:UQ-ARMED:covariate coef}, and the remaining in the supplemental \ref{sup:UQ-ARMED:covariate coef}, and a summary in table \ref{tab:UQ-ARMED:covariate summary}. Between the BNN all, the SWAG-diag lr=0.1, and the ensemble method with a subsampling of 90\%, the ensemble method provided best prediction confidence calibration providing a mean separation between the confidence of correctly and incorrectly predicted subjects of 6\% basis points for the seen test data and 3.6\% basis points for the unseen site data and a highly statistically significant difference between the confidence of the incorrectly predicted and correctly predicted subjects. The Bayesian methods showed a smaller, but still statistically significant different between the correctly predicted and incorrectly predicted confidences with a mean difference of 1.1\% and 0.7\% basis points for the seen and unseen test data respectively. Some of the SWAG methods also provided a statistically significant difference between the confidence of the correctly and incorrectly predicted subjects, for example the SWAG-diag with a lr=0.1 showed a statistically significant difference in confidence of 2.9\% and 2.0\% for the seen and unseen test data respectively. Additionally, for many models including the ensemble model with a sampling of 90\%, the confidence for un-seen test data is lower for both the correctly and incorrectly predicted subjects when compared to the seen test data. This is an favorable characteristic as we expect models to be less confident on out of distribution data, which, by definition, un-seen test data is \emph{iid}.

\begin{figure}[]
  
  \centering
  \includegraphics[width=\textwidth,keepaspectratio]{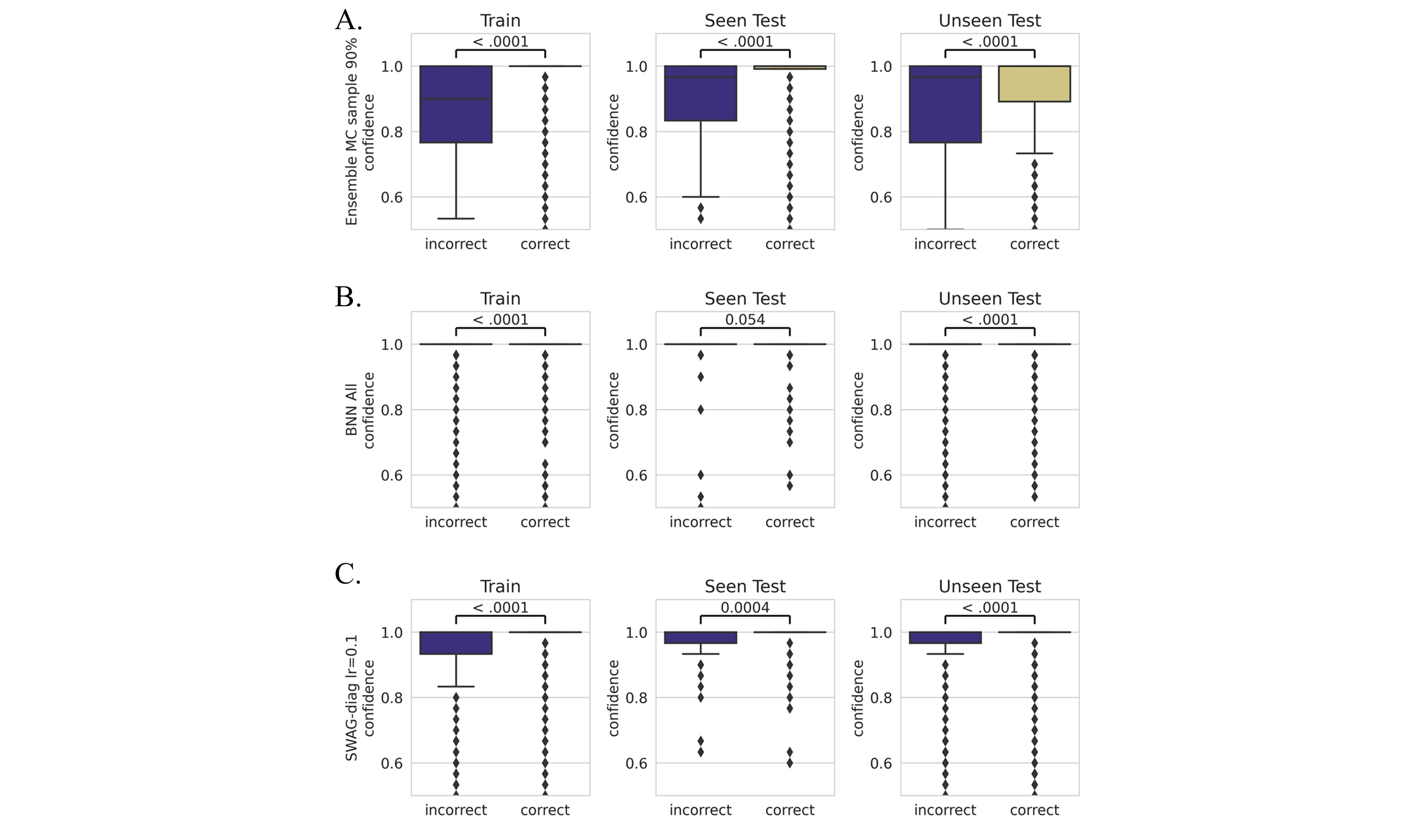}
   \caption[Box plots for comparison of the prediction confidence of correctly and incorrectly predicted subjects per model, for 3 models with high predictive performance.]{
   Box plots for comparison of the prediction confidence of correctly and incorrectly predicted subjects per model, for 3 models with high predictive performance. Models shown are consistent with Fig. \ref{fig:UQ-ARMED:covariate coef}. Confidence is calculated as described in sec \ref{sec:UQ-ARMED:methods:confidence}. The x-axis groups the prediction confidence based on the correctly and incorrectly predicted samples. Columns show the distributions on the training, seen-site test data, and unseen-site test data, left to right, for each model.
    \textbf{A.} Prediction confidence for Ensemble model with a 90\% sampling of the data.
    \textbf{B.} Prediction confidence for BNN model with all layers Bayesian.
    \textbf{C.} Prediction confidence for SWAG-diag model with a 0.1 learning rate.
    Models shown here, are highlighted in green in fig \ref{tab:UQ-ARMED:confidence}.
    Model confidence for all other models can be found in \ref{sup:UQ-ARMED:confidence}
  }
  \label{fig:UQ-ARMED:prediction_confidence}

\end{figure}
% QQ: Please indicate (e.g. with a color) on Fig 4.2 which models (rows) from that table you are showin in the box plots in this figure   

\begin{sidewaystable}[]
\footnotesize
\centering

\begin{tabular}{l|llll|llll|llll}
                               & \multicolumn{4}{l}{Train}                           & \multicolumn{4}{l}{Seen Site}                       & \multicolumn{4}{l}{Unseen site}                     \\
Model                          & Correct & Incorrect & Difference & p              & Correct & Incorrect & Difference & p              & Correct & Incorrect & Difference & p              \\ \hline
Non UQ ARMED                   & 1.000   & 1.000     & 0.000      &                  & 1.000   & 1.000     & 0.000      &                  & 1.000   & 1.000     & 0.000      &                  \\
BNN First                      & 0.997   & 0.984     & 0.013      & \textless{}.0001 & 0.998   & 0.979     & 0.020      & 0.0003           & 0.989   & 0.981     & 0.008      & 0.0004           \\
BNN Last                       & 0.996   & 0.980     & 0.017      & \textless{}.0001 & 0.996   & 0.984     & 0.012      & 0.0325           & 0.993   & 0.990     & 0.002      & 0.168            \\
\textcolor{ForestGreen}{BNN All}& 0.995   & 0.979     & 0.016      & \textless{}.0001 & 0.992   & 0.981     & 0.011      & 0.108            & 0.993   & 0.986     & 0.007      & 0.0002           \\
\textcolor{ForestGreen}{SWAG-diag lr=0.1} & 0.989   & 0.928     & 0.061      & \textless{}.0001 & 0.987   & 0.958     & 0.029      & 0.0007           & 0.971   & 0.951     & 0.020      & \textless{}.0001 \\
SWAG-diag lr=0.01              & 0.965   & 0.925     & 0.040      & \textless{}.0001 & 0.950   & 0.928     & 0.022      & 0.1521           & 0.938   & 0.911     & 0.027      & \textless{}.0001 \\
SWAG-diag lr=0.001             & 0.999   & 0.996     & 0.003      & 0.0023           & 1.000   & 0.996     & 0.004      & 0.019            & 0.998   & 0.997     & 0.001      & 0.4277           \\
SWAG-diag lr=0.0001            & 1.000   & 1.000     & 0.000      & 1.000            & 1.000   & 1.000     & 0.000      & 1.000            & 1.000   & 1.000     & 0.000      & 1.000            \\
SWAG-full lr=0.1               & 0.829   & 0.686     & 0.142      & \textless{}.0001 & 0.817   & 0.694     & 0.123      & \textless{}.0001 & 0.742   & 0.673     & 0.069      & \textless{}.0001 \\
SWAG-full lr=0.01              & 0.870   & 0.699     & 0.171      & \textless{}.0001 & 0.861   & 0.701     & 0.160      & \textless{}.0001 & 0.803   & 0.783     & 0.019      & 0.0022           \\
SWAG-full lr=0.001             & 0.920   & 0.818     & 0.102      & \textless{}.0001 & 0.911   & 0.830     & 0.080      & \textless{}.0001 & 0.877   & 0.858     & 0.019      & 0.0021           \\
SWAG-full lr=0.0001            & 0.926   & 0.844     & 0.082      & \textless{}.0001 & 0.914   & 0.853     & 0.061      & 0.0007           & 0.879   & 0.870     & 0.009      & 0.1475           \\
Ensemble Random Initializations & 0.947   & 0.788     & 0.159      & \textless{}.0001 & 0.927   & 0.789     & 0.138      & \textless{}.0001 & 0.808   & 0.761     & 0.047      & \textless{}.0001 \\
Ensemble MC sample 70\%        & 0.942   & 0.780     & 0.162      & \textless{}.0001 & 0.936   & 0.870     & 0.065      & \textless{}.0001 & 0.887   & 0.864     & 0.023      & \textless{}.0001 \\
Ensemble MC sample 80\%        & 0.953   & 0.827     & 0.127      & \textless{}.0001 & 0.945   & 0.878     & 0.067      & \textless{}.0001 & 0.909   & 0.873     & 0.036      & \textless{}.0001 \\
\textcolor{ForestGreen}{Ensemble MC sample 90\%}        & 0.964   & 0.866     & 0.097      & \textless{}.0001 & 0.954   & 0.894     & 0.060      & 0.0001           & 0.916   & 0.881     & 0.036      & \textless{}.0001 \\
MC Dropout 10\%                & 0.965   & 0.897     & 0.068      & \textless{}.0001 & 0.953   & 0.896     & 0.057      & \textless{}.0001 & 0.938   & 0.924     & 0.014      & 0.0015           \\
MC Dropout 20\%                & 0.938   & 0.854     & 0.084      & \textless{}.0001 & 0.920   & 0.891     & 0.029      & 0.0642           & 0.918   & 0.908     & 0.010      & 0.0336           \\
MC Dropout 30\%                & 0.938   & 0.852     & 0.086      & \textless{}.0001 & 0.914   & 0.870     & 0.044      & 0.0041           & 0.898   & 0.882     & 0.017      & 0.0006           \\
MC Dropout 40\%                & 0.924   & 0.862     & 0.062      & \textless{}.0001 & 0.898   & 0.892     & 0.006      & 0.6528           & 0.912   & 0.898     & 0.014      & 0.0005           \\
MC Dropout 50\%                & 0.929   & 0.871     & 0.058      & \textless{}.0001 & 0.917   & 0.892     & 0.025      & 0.0172           & 0.920   & 0.900     & 0.020      & \textless{}.0001
\end{tabular}

\caption[Mean prediction confidence for each model, for the correctly (correct) and incorrectly (incorrect) predicted MCI conversion.]{
Each model's mean prediction confidence for the correctly (correct) and incorrectly (incorrect) predicted MCI conversion, as calculated in sec \ref{sec:UQ-ARMED:methods:confidence}. The difference and p columns, are the difference in confidence between the correctly and incorrectly predicted subject and associated p-value calculated using a 2 sided students t-test, respectively. The larger the difference, the more accurate the confidence in prediction. The models in green are displayed in Fig. \ref{fig:UQ-ARMED:prediction_confidence}.
}
\label{tab:UQ-ARMED:confidence}

\end{sidewaystable}
% QQ: to make this table much more readable, add a vertical bar (line) immediately to the right of both p-value columns  

\section{Time Comparison}
\begin{figure}[]
  
  \centering
  \includegraphics[width=\textwidth,keepaspectratio]{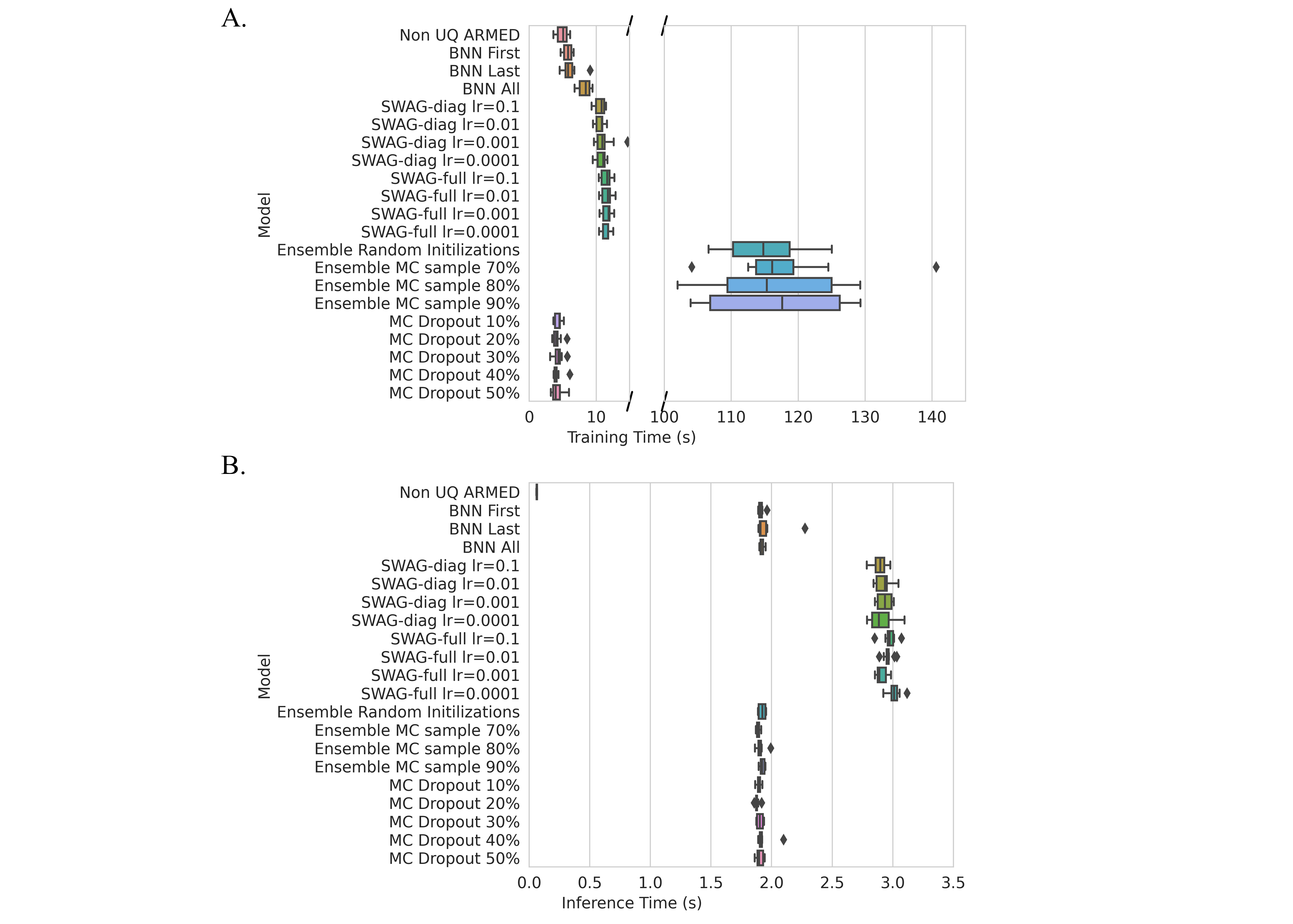}
  \caption[Training and inference time for the non-UQ ARMED and the different UQ-ARMED models. ]{
  Training and inference time for the non-UQ ARMED and the different UQ-ARMED models. 
  \textbf{A.} Training time for each model
  \textbf{B.} Inference time for each model, with 30 samples from the models weights.  
  }
  \label{fig:UQ-ARMED:TrainingTime}

\end{figure}

Overall, the training time for the ensemble methods take about 30 times longer to train than the other models, which is as expected given that the ensembles require training 30 independent models \ref{fig:UQ-ARMED:TrainingTime}. However, this training task is an embarrassingly parallel problem, meaning that the models could be trained independently in parallel, and as many ML practitioners have access to such parallel compute infrastructure this may not be a barrier. The inference time for all Bayesian deep learning models, excluding the SWAG models, is about 30 times longer than the regular ARMED models. This is also expected as each model is required to produce 30 predictions to estimate the posterior distribution. The SWAG models required about 50 times the inference time. The additional time is due to the sampling of the co-variance matrices. However, after the UQ models have been trained, each inference for all UQ models is independent, and thus could also be paralleled, effectively reducing the inference by 30x. For all models except SWAG this would make the inference time equivalent to the non-UQ ARMED model, given enough compute. Additionally, with a minor update to SWAG, the sampling of the variance/co-variance weight matrix could be completed and saved. This would make inference time for SWAG equivalent to the other UQ methods. While it would also be possible to increase inference time for the non-UQ ARMED model with \emph{faster} compute, inference through a single model is sequential, and not readily parallelizable, therefore keeping all timing metrics relative and additional compute would still compensate for the additional training and inference time. 

\begin{table}[]

\centering

    \begin{tabular}{lll}
    Model                          & Training Time (s) & Inference Unseen Site Test (s) \\ \hline
    BNN All                        & 8.23              & 1.92                           \\
    BNN First                      & 5.73              & 1.91                           \\
    BNN Last                       & 6.06              & 6.70                           \\
    Ensemble MC sample 70\%        & 118.01            & 1.89                           \\
    Ensemble MC sample 80\%        & 119.47            & 1.91                           \\
    Ensemble MC sample 90\%        & 119.57            & 1.92                           \\
    Ensemble Random Initializations & 114.76            & 1.92                           \\
    MC Dropout 10\%                & 4.34              & 1.89                           \\
    MC Dropout 20\%                & 4.14              & 1.88                           \\
    MC Dropout 30\%                & 4.32              & 1.90                           \\
    MC Dropout 40\%                & 4.14              & 1.93                           \\
    MC Dropout 50\%                & 4.19              & 1.90                           \\
    Non UQ ARMED                   & 4.94              & 0.06                           \\
    SWAG-diag lr=0.0001            & 10.76             & 2.91                           \\
    SWAG-diag lr=0.001             & 11.15             & 2.93                           \\
    SWAG-diag lr=0.01              & 10.59             & 2.92                           \\
    SWAG-diag lr=0.1               & 10.56             & 2.89                           \\
    SWAG-full lr=0.0001            & 11.52             & 3.01                           \\
    SWAG-full lr=0.001             & 11.61             & 2.91                           \\
    SWAG-full lr=0.01              & 11.59             & 2.96                           \\
    SWAG-full lr=0.1               & 11.47             & 2.97                          
    \end{tabular}

\caption[Mean training and inference time in seconds for each model.]{
    Mean training and inference time in seconds for each model.
}
\label{tab:UQ-ARMED:time}

\end{table}

\chapter{Discussion}
% QQ: Discussion always precedes conclusions.  Disc is to bring up speculative points -- your interpretation of the results and speculations as to why you got the results you did.   
%-- you should also include brief list of limitations
% QQ: Make sure conclusions includes the final recommendation and it ends on positive note of what you did.

Based on these results, we recommend the ensemble UQ methods. Ensembling UQ performed well across multiple metrics including: prediction performance, probe de-weighting and covariate significance estimation, and confidence calibration. Ensembles were also shown to be the most easy to configure as they demonstrate they are less sensitive to changes in the hyperparameter compared to the other methods. For example, all ensemble methods using sub-sampling performed well compared to the non-UQ ARMED models and provided non-statistically significant FE covariate coefficents for the synthetic probes. Whereas the results for the SWAG models was highly dependent on the learning rate. The largest downside of the ensemble models is the training time, as they took about 30 times longer to train when compared to the other UQ methods tested in this work. Furthermore, the additional training time scales linearly with the number of samples from the posterior required. E.g. for 100 samples, the ensemble methods would take about 100x longer to train than the other approaches. Whereas the training time for BNN's, SWAG, and MC dropout models does not depend on the number posterior samples. Additionally, the ensemble methods require the a priori selection of the number of samples from the prior before training. All other methods allow you to sample from the prior, a theoretically, infinite number of times without any additional training.

Many of the Bayesian networks performed similarly to the standard ARMED model \ref{tab:UQ-ARMED:performance}, including the ensemble method using sub-sampling, the BNNs, and some of the SWAG models. One notable decrease in performance is balanced accuracy of the ensemble approach with random initializations. One possibility is, because the model started in different random initializations, each ensemble method found a different local minima, a small subset of which is not as optimal, which would cause the large variance in performance. Whereas the ensembles methods that used the same initialization likely sampled from relatively more similar minima (interestingly, similar to SWAG). However, this drop in performance was successfully captured by the p-value of the model fit, Table \ref{tab:UQ-ARMED:performance}.

Based on the 95\% CI, the MC models provided the worst performance, and was the only method where all models performed statistically significantly worse than the non-UQ ARMED model. As per the experimental design, the hyperparameters of the conventional models are keep consistent to provide the fairest comparison to the standard ARMED model. However, as dropout essentially removes model weights, MC models may benefit from increasing the number of neurons per layer in proportion to the dropout rate. This would however require additional optimization, would introduce complexity, be computationally costly, and obscure these results.

When comparing the sensitivity of the models UQ to the tested hyperparameters, the subsampling ensemble methods seemed most robust to change in the tested hyperparameter which is a desirable trait. However, with a smaller dataset, subsampling may become problematic as the ensemble modes may not have enough data to train on. Further experimentation would be required to better quantify this effect. For the SWAG models, every metric measured in this work, performance, statistical significance of model fit and covariate coefficient, and prediction confidence varied relatively largely, compared to the other UQ models, based learning rate of SWAG while estimating the co-variance matrix. This suggests that it is critical to carefully optimize the learning rate used when sampling the weights when using SWAG. This is expected, as distance traveled in the loss space is directly proportional to the learning rate, in turn directly effecting the sampled weights that moderates the estimated variance/co-variance matrix, which is also described in the manuscript \cite{Maddox.2019}. For the MC dropout models we saw a general trend in a decrease in performance as the drop out rate increases. 

In this work we provide 2 reasonable approaches to estimate the covariate coefficient based on the model gradient. The $ave(\frac{\partial y}{\partial x})$ approach samples the covariate coefficient across the feature space rather than at a single point, which may lead to a better approximation. However, it is perhaps more susceptible to outliers or biasing the covariate coefficient to subjects that are in low density areas of covariate space. $\frac{\partial y}{\partial \Bar{x}}$ provides the gradient at the center of the feature space, where the is most support, but is less flexible in accounting for large variations of the gradient over the input space. We acknowledge the advantages of both approaches and provide the results for the $\frac{\partial y}{\partial \Bar{x}}$in the supplemental.

Lastly, our work has the following limitations. To provide uncertainty quantification, we employ methods that model epistemic uncertainty, that models the model uncertainty of the weights given the data $p(w|D)$, rather than aleatoric uncertainty. Future work that models both epistemic and aleatoric uncertainty may increase the accuracy of the uncertainty quantification. Additionally, this work does not estimate uncertainty from sources of variation not included in the experiment (e.g additional unseen sites), and thus the uncertainties are lower bounds. However, in our AD experiment we do have 34 un-seen sites, we would argue that the random effect is sufficiently sampled such that the upper and lower bounds on uncertainty are likely similar. This may not be true with data sets with significantly less numbers of groupings in the random effects. Another limitation of this work is the application to a single dataset. While ADNI provides an ideal dataset to test the UQ-ARMED models on due to the large number of sites, future work should aim to compare these results on other data sets with known random effects. 

\chapter{Conclusions}
This work demonstrates the ability to produce readily interpretable statistical metrics for model fit, fixed effects covariance coefficients, and prediction confidence. Importantly, this work compares 4 suitable and commonly applied epistemic UQ approaches, BNN, SWAG, MC dropout, and ensemble approaches in their ability to calculate these statistical metrics for the ARMED MEDL models. In our experiment, not only do the UQ methods provide these benefits, but several UQ methods maintain the high performance of the original ARMED method, some even provide a modest (but not statistically significant) performance improvement. The ensemble models, especially the ensemble method with a 90\% subsampling, performed well across all metrics we tested with (1) high performance that was comparable to the non-UQ ARMED model, (2) properly deweights the confounds probes and assigns them statistically insignificant p-values, (3) attains relatively high calibration of the output prediction confidence. The MC dropout models showed the lowest performance, and failed to provide non-statistically significant fixed effects covariate coefficients. The SWAG model’s performance was dependent on the learning rate. Specifically, the models with low learning rate underestimated the fixed effect covariate coefficient uncertainty providing very small standard errors causing very significant p-values for all covariate coefficients including for the synthetic probes. Lastly, The BNNs also performed reasonably well, showing good model performance, and almost achieved providing statistically insignificant p-values for the synthetic probe covariate coefficients (depending on your cut off), but did not perform as well as the ensemble approaches for both covariate coefficient statistical significance, or model prediction confidence. The largest potential downside to the ensemble approaches is the increased training time, however as discussed in results, the balance between wall clock time and available computational resources could be achieved through parallelization. Additionally, in many instances the models inference time is more important than training time, for which the ensemble approaches are tied as the fastest of the UQ methods. Based on these results, the ensemble approaches, especially with a subsampling of 90\%, provided the best all-round performance for prediction and uncertainty estimation, and achieved our goals to provide statistical significance for model fit, statistical significance covariate coefficients, and confidence in prediction, while maintaining the baseline performance of MEDL using ARMED.

\chapter{Supplemental Material}
\label{appendix:UQ-ARMED}
%%%%%%%%%%%%%%%%%%%%%%%%%%%%%%%%%%%%%%%%%%%%%%%%%%%%%%%%%%%%%%%%%%%%%%%%%
%model hyperparameter table
%%%%%%%%%%%%%%%%%%%%%%%%%%%%%%%%%%%%%%%%%%%%%%%%%%%%%%%%%%%%%%%%%%%%%%%%%
\begin{table}[]
    \begin{tabular}{ll}
    Hyperparameter                                             & Value                     \\ \hline
    Number of hidden layers in conventional model              & 4                         \\
    Number of neurons in each layer for the conventional model & 4,4,4,4                     \\
    Optimizer                                                  & ADAM                      \\
    Random effects network type                                & Linear                    \\
    Classification loss (FE, ME)                               & Binary Cross Entropy      \\
    Classificatoin loss  (Adversarial)                         & Categorical Cross Entropy
        
    \end{tabular}
    \caption[Hyperparameters for non-UQ ARMED model.]{
    Hyperparameters for non-UQ ARMED model. These values are consistent with \cite{Nguyen.2022b}
    }
    \label{sup:UQ-ARMED:HP}
\end{table}
\clearpage

%%%%%%%%%%%%%%%%%%%%%%%%%%%%%%%%%%%%%%%%%%%%%%%%%%%%%%%%%%%%%%%%%%%%%%%%%
%%%covariate coef est for all models using same method in main
%%%%%%%%%%%%%%%%%%%%%%%%%%%%%%%%%%%%%%%%%%%%%%%%%%%%%%%%%%%%%%%%%%%%%%%%%
%There are a lot of figures, so I split them so it can span multiple pages.
%not sure if this is the best way of achieving this=, but it does seem to work
%add the caption to the last figure
\begin{figure}[]
  
  \centering
  \includegraphics[width=\textwidth,keepaspectratio]{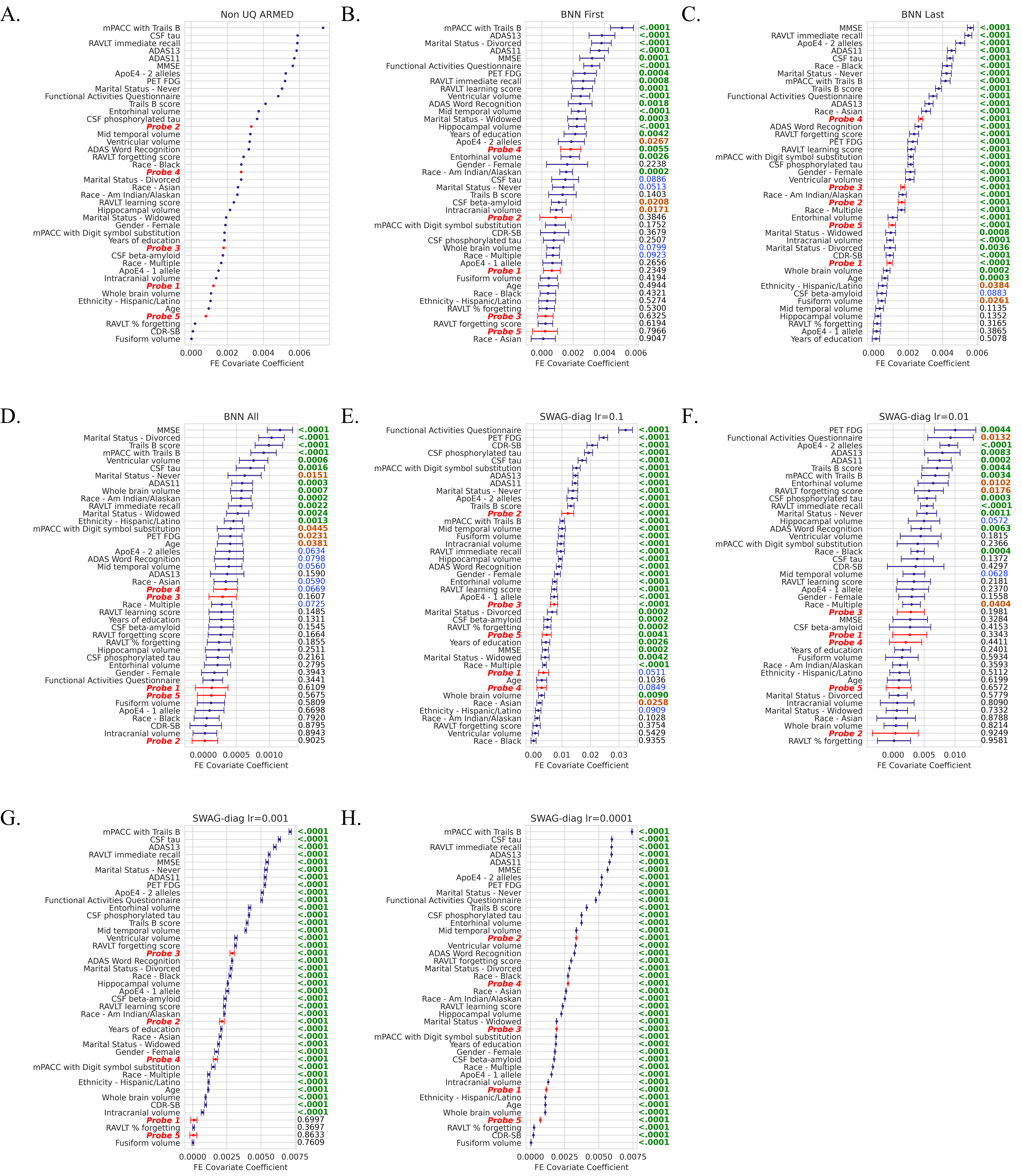}

\end{figure}

\clearpage

\begin{figure}[]
  
  \centering
  \includegraphics[width=\textwidth,keepaspectratio]{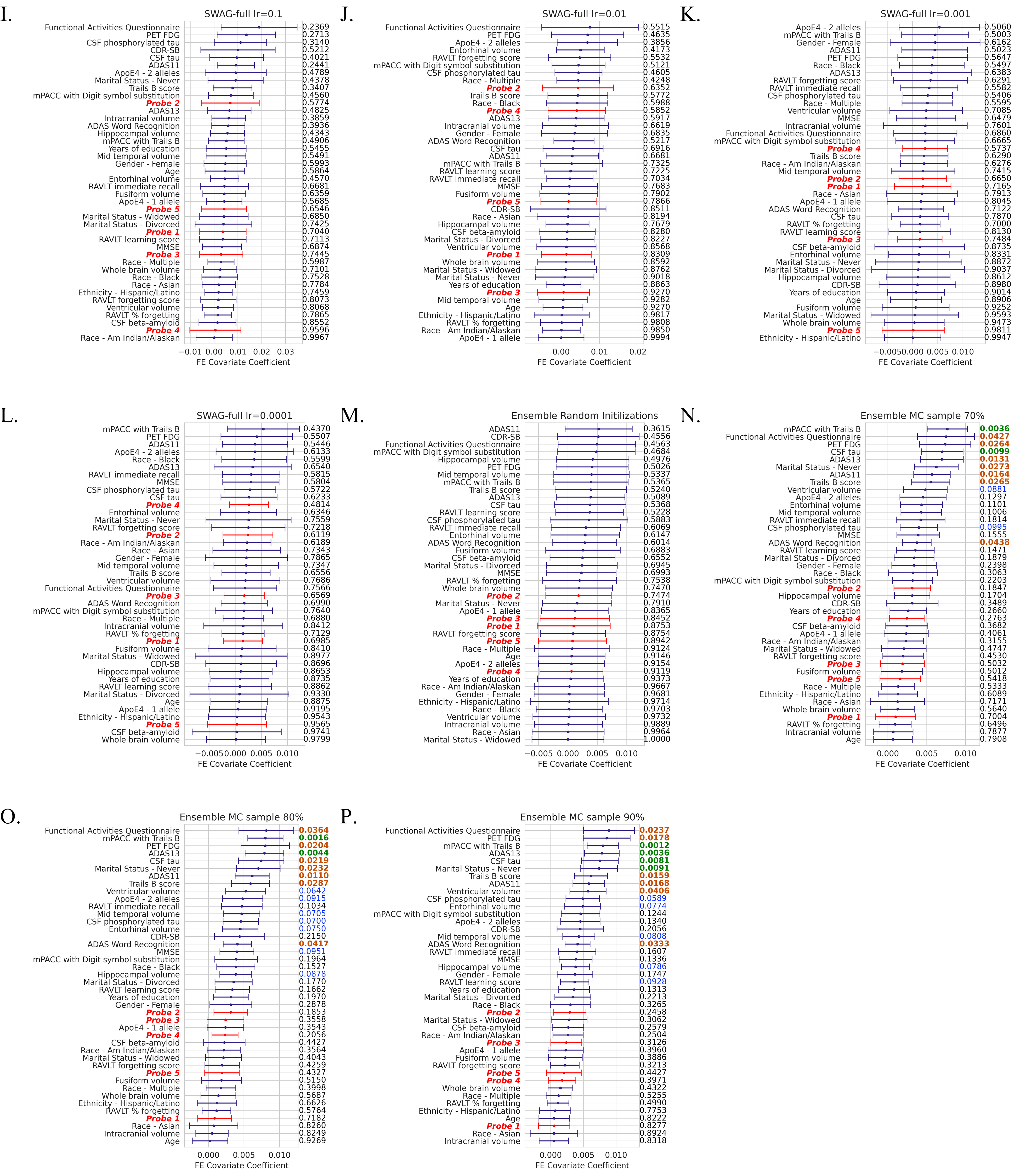}

\end{figure}

\clearpage

\begin{figure}[]
  
  \centering
  \includegraphics[width=\textwidth,keepaspectratio]{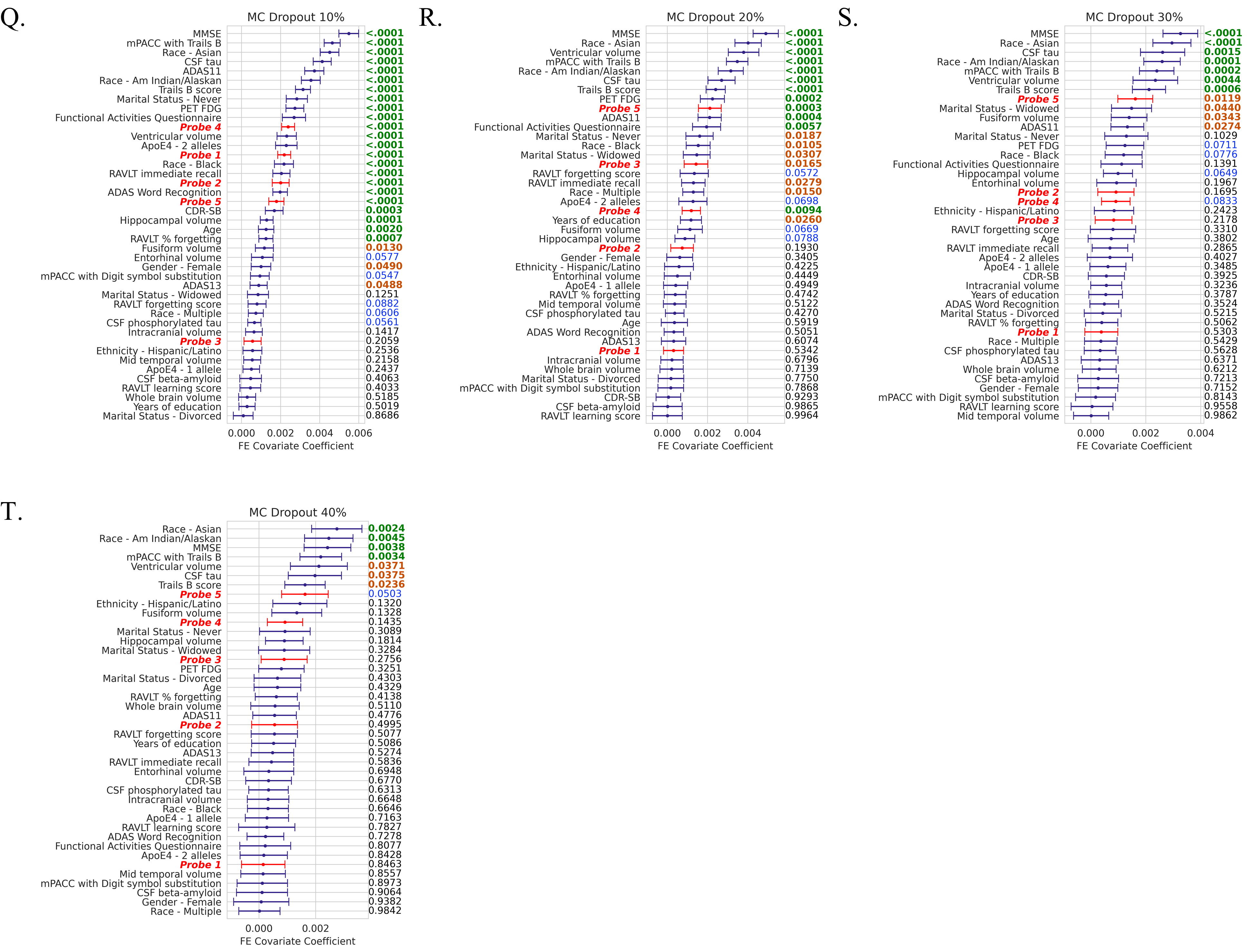}
  \caption[Coefficient covariate estimation and calculated uncertainty for all models.]{
  Coefficient covariate estimation and calculated uncertainty for all models. Each plot shows the mean, and 95CI for each estimated covariate coefficient. Specific model indicated at the top of each plot. Organized in the same order as the performance plots in Fig. \ref{fig:UQ-ARMED:performance}.
  }
  \label{sup:UQ-ARMED:covariate coef}

\end{figure}
\clearpage

%%%%%%%%%%%%%%%%%%%%%%%%%%%%%%%%%%%%%%%%%%%%%%%%%%%%%%%%%%%%%%%%%%%%%%%%%
%%%covariate coef est for all models using dy/dxbar. Probably add paragraph to describe these results
%%%%%%%%%%%%%%%%%%%%%%%%%%%%%%%%%%%%%%%%%%%%%%%%%%%%%%%%%%%%%%%%%%%%%%%%%

\section{Coefficient covariate estimation using $\frac{\partial y}{\partial \Bar{x}}$}
As described in \ref{sec:UQ-ARMED:methods:covariate coef} there are two reasonable approaches for estimating the covariate coefficient calculation for the non-linear models.

\clearpage
\begin{figure}[]
  
  \centering
  \includegraphics[width=\textwidth,keepaspectratio]{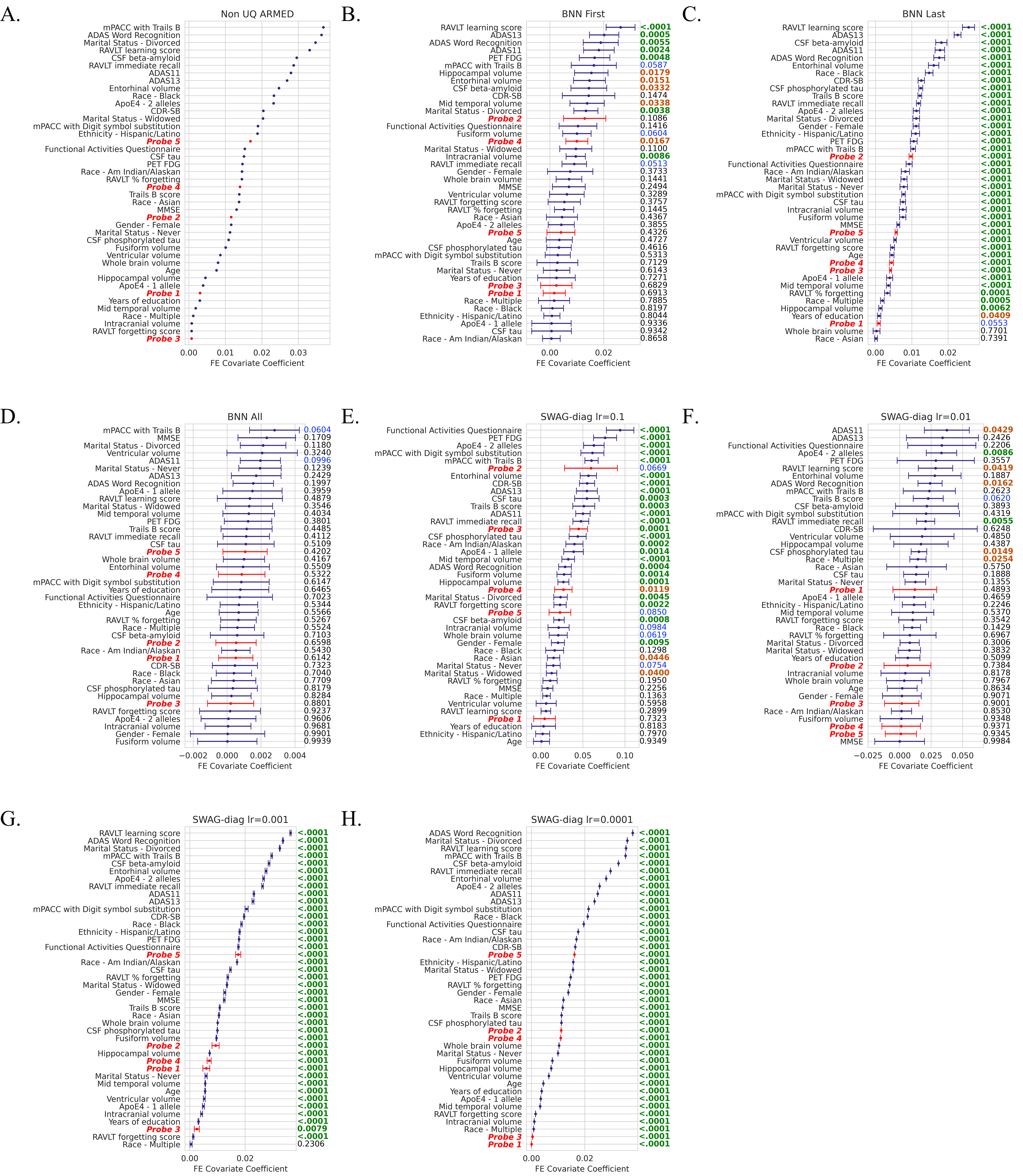}

\end{figure}

\clearpage

\begin{figure}[]
  
  \centering
  \includegraphics[width=\textwidth,keepaspectratio]{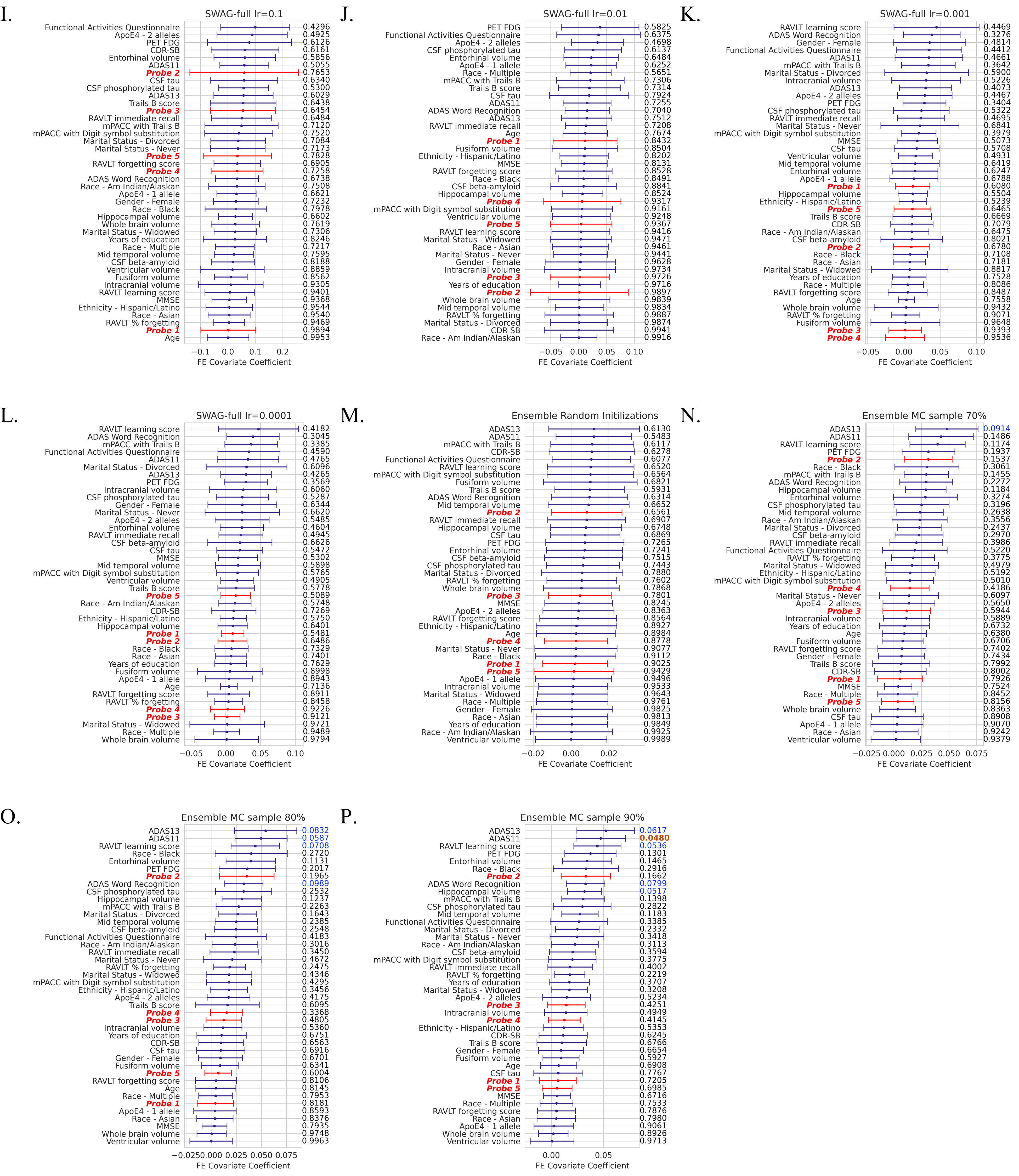}

\end{figure}

\clearpage

\begin{figure}[]
  
  \centering
  \includegraphics[width=\textwidth,keepaspectratio]{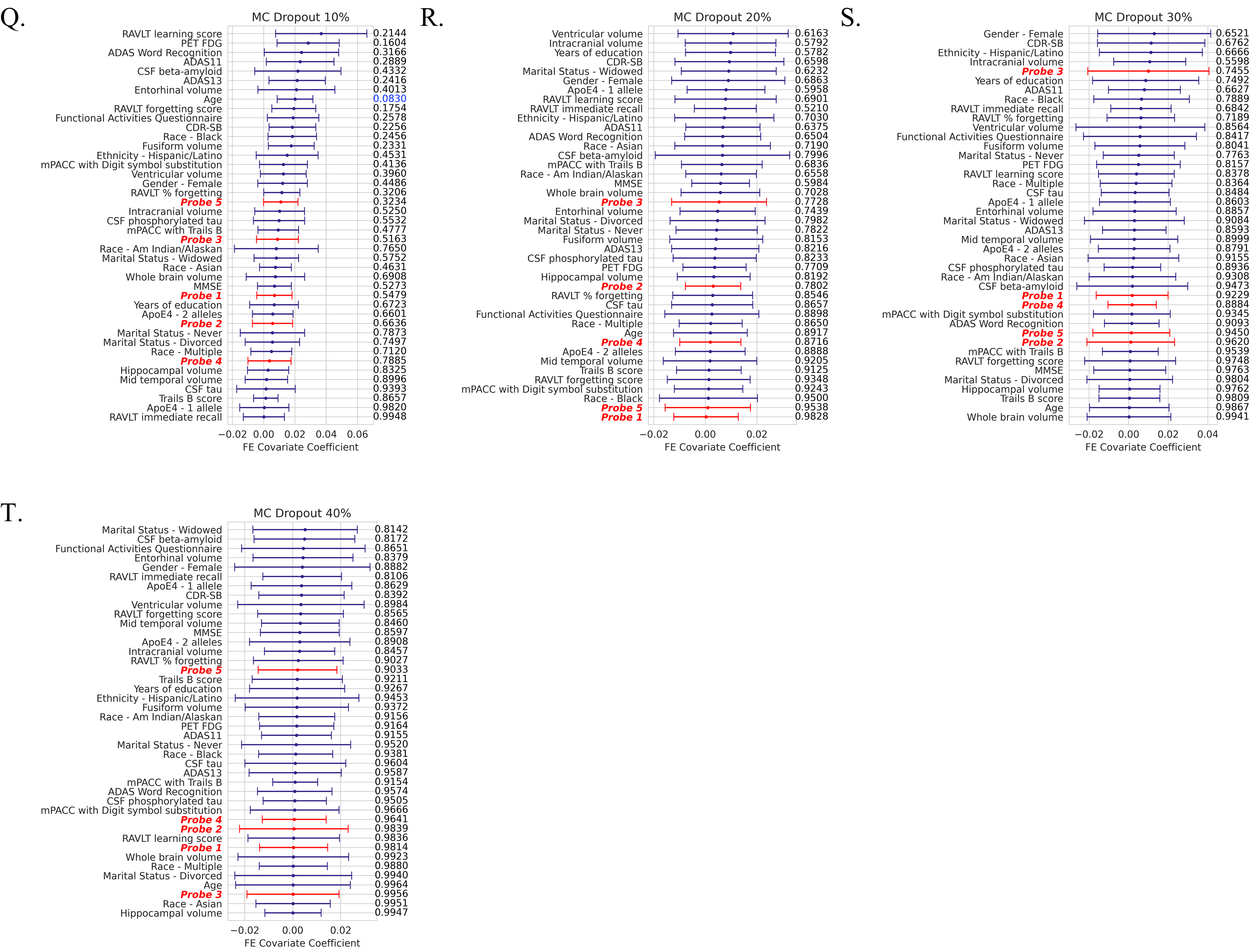}
  \caption[Coefficient covariate estimation and calculated uncertainty for all models using $\frac{\partial y}{\partial \Bar{x}}$.]{
    Coefficient covariate estimation and calculated uncertainty for all models using $\frac{\partial y}{\partial \Bar{x}}$. Each plot shows the mean, and 95CI for each estimated covariate coefficient. Specific model indicated at the top of each plot. Organized in the same order as the performance plots in Fig. \ref{fig:UQ-ARMED:performance}.
  }
  \label{sup:fig:UQ-ARMED:covariate coef dy/dxbar}

\end{figure}

%%%%%%%%%%%%%%%%%%%%%%%%%%%%%%%%%%%%%%%%%%%%%%%%%%%%%%%%%%%%%%%%%%%%%%%%%
%%%prediction confidence for each model
%%%%%%%%%%%%%%%%%%%%%%%%%%%%%%%%%%%%%%%%%%%%%%%%%%%%%%%%%%%%%%%%%%%%%%%%%

\begin{figure}[]
  
  \centering
  \includegraphics[width=\textwidth,keepaspectratio]{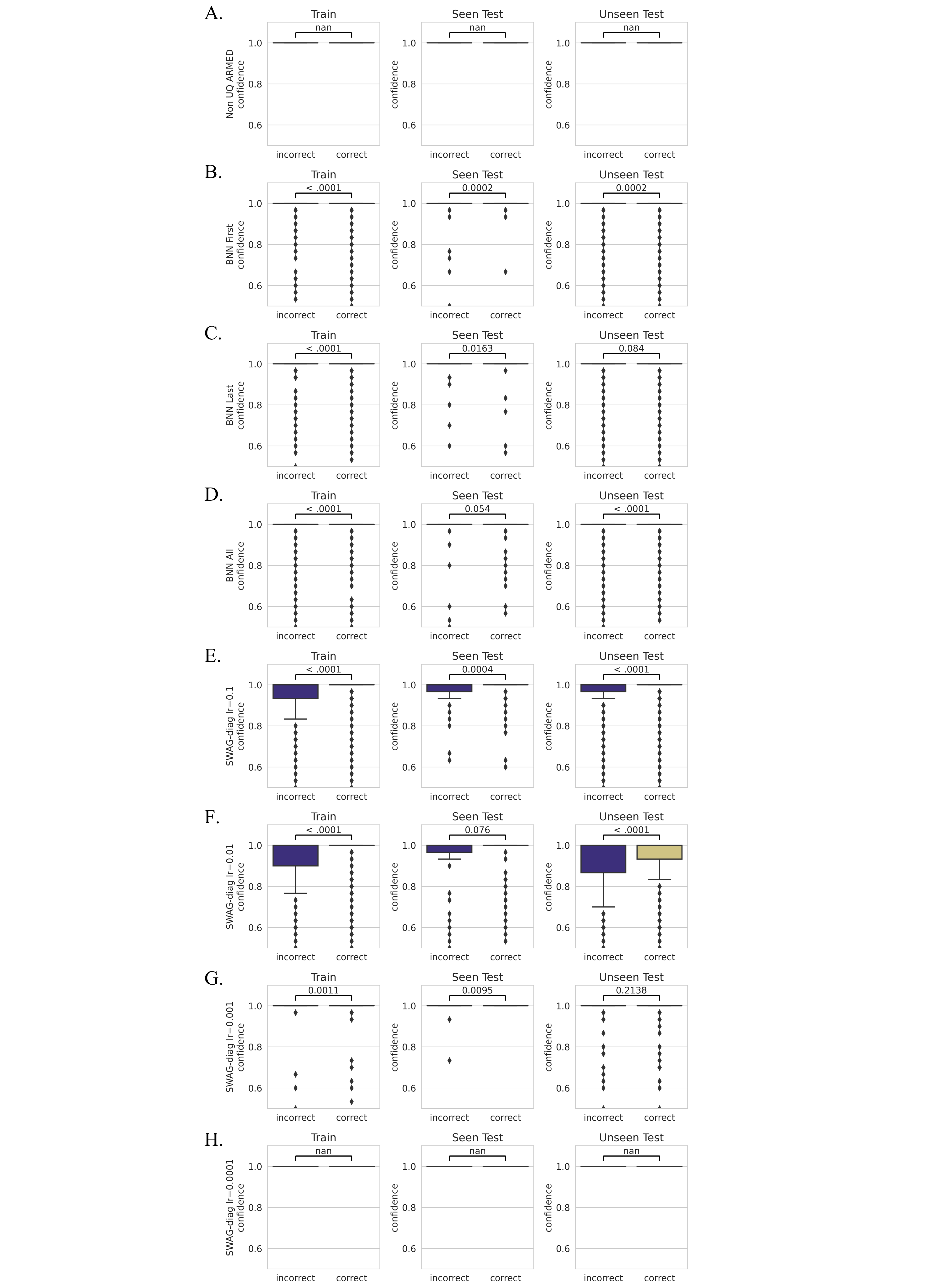}

\end{figure}

\clearpage

\begin{figure}[]
  
  \centering
  \includegraphics[width=\textwidth,keepaspectratio]{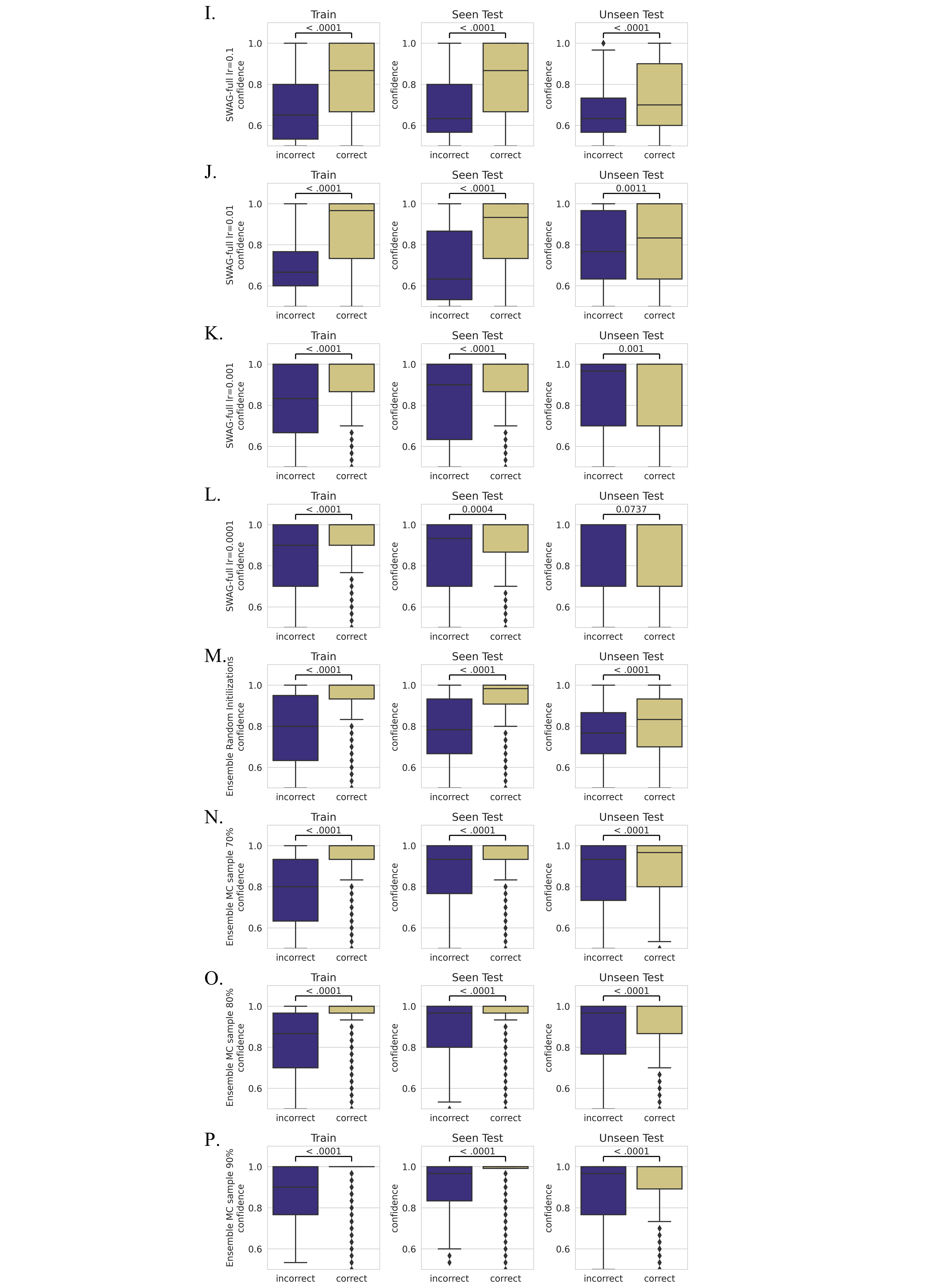}

\end{figure}

\clearpage

\begin{figure}[]
  
  \centering
  \includegraphics[width=\textwidth,keepaspectratio]{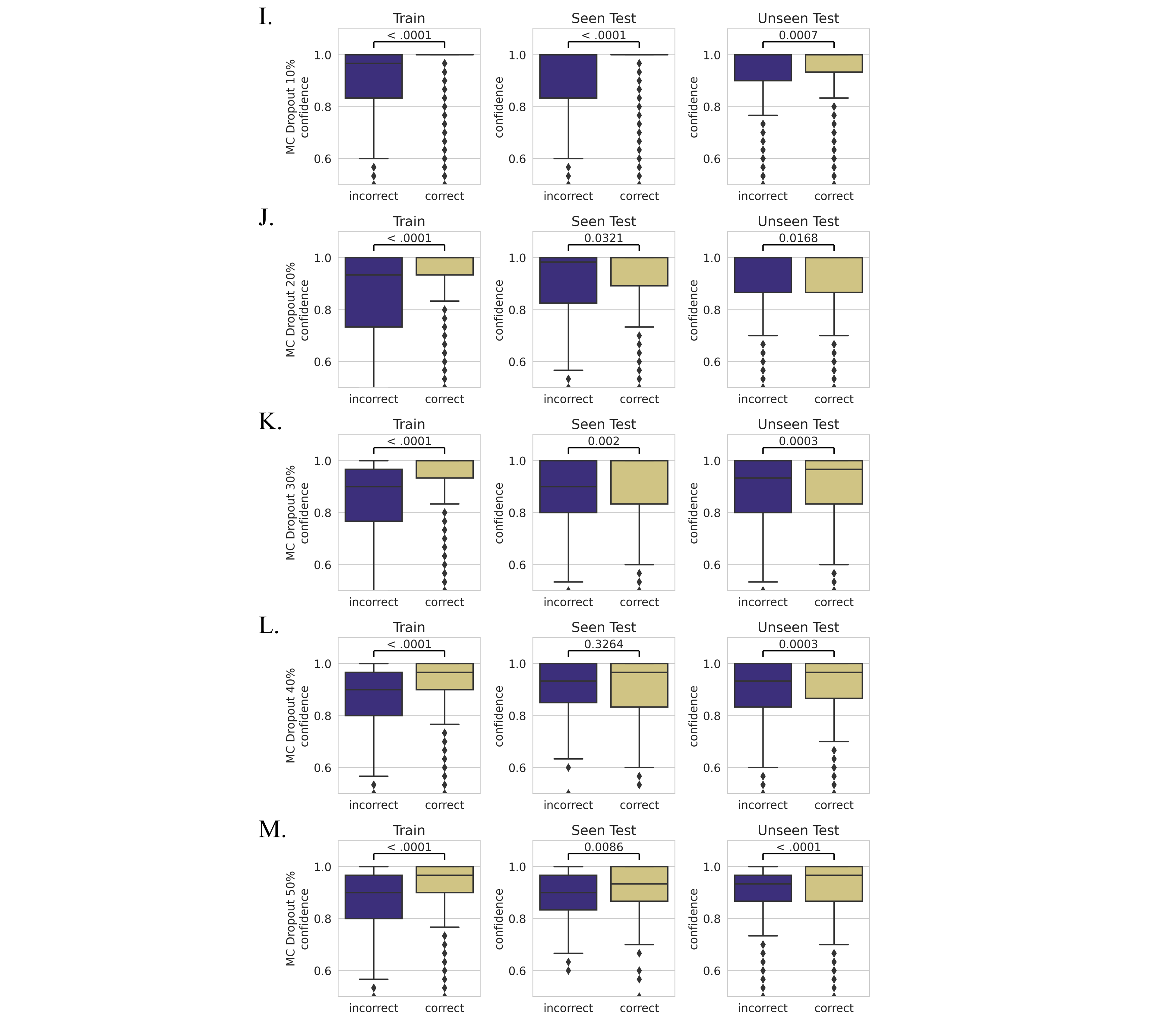}
   \caption[Box plots for comparison of the prediction confidence of correctly and incorrectly predicted subjects per model, for all models]{
   Box plots for comparison of the prediction confidence of correctly and incorrectly predicted subjects per model, for all models with high predictive performance. Confidence is calculated as described in sec \ref{sec:UQ-ARMED:methods:confidence}. The x-axis groups the prediction confidence based on the correctly and incorrectly predicted samples. Columns show the distributions on the training, seen-site test data, and unseen-site test data, left to right, for each model.}
  \label{sup:UQ-ARMED:confidence}

\end{figure}

% Begin the bibliography:

%\begin{thesisbib}  % <--- THIS LINE IS REQUIRED!
\bibliography{bibliography.bib}

\begin{thebibliography}{16}
\providecommand{\natexlab}[1]{#1}
\providecommand{\url}[1]{\texttt{#1}}
\expandafter\ifx\csname urlstyle\endcsname\relax
  \providecommand{\doi}[1]{doi: #1}\else
  \providecommand{\doi}{doi: \begingroup \urlstyle{rm}\Url}\fi

\bibitem[Abdullah et~al.(2022)Abdullah, Hassan, and Mustafa]{Abdullah.2022}
Abdullah~A. Abdullah, Masoud~M. Hassan, and Yaseen~T. Mustafa.
\newblock A review on bayesian deep learning in healthcare: Applications and
  challenges.
\newblock \emph{IEEE Access}, 10:\penalty0 36538--36562, 2022.
\newblock \doi{10.1109/ACCESS.2022.3163384}.

\bibitem[Combrisson and Jerbi(2015)]{Combrisson.2015}
Etienne Combrisson and Karim Jerbi.
\newblock Exceeding chance level by chance: The caveat of theoretical chance
  levels in brain signal classification and statistical assessment of decoding
  accuracy.
\newblock \emph{Journal of neuroscience methods}, 250:\penalty0 126--136, 2015.
\newblock ISSN 0165-0270.
\newblock \doi{10.1016/j.jneumeth.2015.01.010}.

\bibitem[Gal and Ghahramani(2016{\natexlab{a}})]{Gal.2016}
Yarin Gal and Zoubin Ghahramani.
\newblock Dropout as a bayesian approximation: Representing model uncertainty
  in deep learning.
\newblock \emph{Proceedings of Machine Learning Research}, pages 1--10,
  2016{\natexlab{a}}.
\newblock URL \url{http://proceedings.mlr.press/v48/gal16.pdf}.

\bibitem[Gal and Ghahramani(2016{\natexlab{b}})]{Gal.2016b}
Yarin Gal and Zoubin Ghahramani.
\newblock A theoretically grounded application of dropout in recurrent neural
  networks.
\newblock \emph{Advances in Neural Information Processing Systems (NeurIPS)},
  29:\penalty0 1--9, 2016{\natexlab{b}}.
\newblock URL
  \url{https://proceedings.neurips.cc/paper/2016/file/076a0c97d09cf1a0ec3e19c7f2529f2b-Paper.pdf}.

\bibitem[Hau{\ss}mann et~al.(2020)Hau{\ss}mann, Hamprecht, and
  Kandemir]{Haubmann.2020}
Manuel Hau{\ss}mann, Fred~A. Hamprecht, and Melih Kandemir.
\newblock Sampling-free variational inference of bayesian neural networks by
  variance backpropagation.
\newblock \emph{Proceedings of Machine Learning Research}, 115:\penalty0
  563--573, 2020.
\newblock URL \url{https://proceedings.mlr.press/v115/haussmann20a.html}.

\bibitem[Hornik et~al.(1989)Hornik, Stinchcombe, and White]{Hornik.1989}
Kurt Hornik, Maxwell Stinchcombe, and Halbert White.
\newblock Multilayer feedforward networks are universal approximators.
\newblock \emph{Neural Networks}, 2\penalty0 (5):\penalty0 359--366, 1989.
\newblock ISSN 08936080.
\newblock \doi{10.1016/0893-6080(89)90020-8}.

\bibitem[Maddox et~al.(2019)Maddox, Garipov, Izmailov, Vetrov, and
  Wilson]{Maddox.2019}
Wesley Maddox, Timur Garipov, Pavel Izmailov, Dmitry Vetrov, and Andrew~Gordon
  Wilson.
\newblock A simple baseline for bayesian uncertainty in deep learning.
\newblock \emph{ArXiv preprint}, pages 1--25, 2019.
\newblock URL \url{http://arxiv.org/pdf/1902.02476v2}.

\bibitem[Mena et~al.(2022)Mena, Pujol, and Vitri{\`a}]{Mena.2022}
Jos{\'e} Mena, Oriol Pujol, and Jordi Vitri{\`a}.
\newblock A survey on uncertainty estimation in deep learning classification
  systems from a bayesian perspective.
\newblock \emph{ACM Computing Surveys (CSUR)}, 54:\penalty0 1--35, 2022.

\bibitem[Monarch and Manning(2021)]{Monarch.2021}
Robert Monarch and Christopher~D. Manning.
\newblock \emph{Human-in-the-loop machine learning: Active learning and
  annotationfor human-centered AI}.
\newblock {Manning Publications}, Shelter Island, New York, 1st edition
  edition, 2021.
\newblock ISBN 9781617296741.
\newblock URL
  \url{https://learning.oreilly.com/library/view/-/9781617296741/?ar}.

\bibitem[Nguyen et~al.(2022)Nguyen, {\bf Treacher, Alex H.}, and
  Montillo]{Nguyen.2022b}
Kevin~P. Nguyen, {\bf Treacher, Alex H.}, and Albert Montillo.
\newblock Adversarially-regularized mixed effects deep learning ({ARMED})
  models for improved interpretability, performance, and generalization on
  clustered data.
\newblock \emph{IEEE transactions on pattern analysis and machine intelligence
  (in press)}, 2022.
\newblock URL \url{https://arxiv.org/pdf/2202.11783}.

\bibitem[Nowozin et~al.(2016)Nowozin, Cseke, and Tomioka]{Nowozin.2016}
Sebastian Nowozin, Botond Cseke, and Ryota Tomioka.
\newblock f-gan: Training generative neural samplers using variational
  divergence minimization.
\newblock \emph{NeurIPS}, pages 1--9, 2016.
\newblock URL \url{https://arxiv.org/pdf/1606.00709}.

\bibitem[Simchoni and Rosset(2021)]{Simchoni.2021}
Giora Simchoni and Saharon Rosset.
\newblock Using random effects to account for high-cardinality categorical
  features and repeated measures in deep neural networks.
\newblock \emph{Advances in Neural Information Processing Systems (NeurIPS)},
  34:\penalty0 25111--25122, 2021.
\newblock URL
  \url{https://proceedings.neurips.cc/paper/2021/file/d35b05a832e2bb91f110d54e34e2da79-Paper.pdf}.

\bibitem[Srivastava et~al.(2014)Srivastava, Hinton, Krizhevsky, Sutskever, and
  Salakhutdinov]{Srivastava.2014}
Nitish Srivastava, Geoffrey Hinton, Alex Krizhevsky, Ilya Sutskever, and Ruslan
  Salakhutdinov.
\newblock Dropout: A simple way to prevent neural networks from overfitting.
\newblock \emph{Journal of Machine Learning Research}, 15\penalty0
  (56):\penalty0 1929--1958, 2014.
\newblock URL \url{http://jmlr.org/papers/v15/srivastava14a.html}.

\bibitem[Wagner(1982)]{Wagner.1982}
Clifford~H. Wagner.
\newblock Simpson's paradox in real life.
\newblock \emph{The American Statistician}, 36:\penalty0 46--48, 1982.
\newblock ISSN 00031305.
\newblock \doi{10.2307/2684093}.

\bibitem[Wilson and Izmailov(2020)]{Wilson.2020}
Andrew~G. Wilson and Pavel Izmailov.
\newblock Bayesian deep learning and a probabilistic perspective of
  generalization.
\newblock \emph{Advances in Neural Information Processing Systems (NeurIPS)},
  33:\penalty0 4697--4708, 2020.
\newblock URL
  \url{https://proceedings.neurips.cc/paper/2020/file/322f62469c5e3c7dc3e58f5a4d1ea399-Paper.pdf}.

\bibitem[Xiong et~al.(2019)Xiong, Kim, and Singh]{Xiong.2019}
Yunyang Xiong, Hyunwoo~J. Kim, and Vikas Singh.
\newblock Mixed effects neural networks (menets) with applications to gaze
  estimation.
\newblock \emph{Conference on Computer Vision and Pattern Recognition (CVPR)},
  pages 7743--7752, 2019.

\end{thebibliography}
%\end{thesisbib}  % <-- THIS LINE IS REQUIRED!

\end{document}